\documentclass[10pt,twocolumn,letterpaper]{article}

\usepackage[pagenumbers]{cvpr}     
\definecolor{cvprblue}{rgb}{0.21,0.49,0.74}
\usepackage[pagebackref,breaklinks,colorlinks,allcolors=cvprblue]{hyperref}
\usepackage{multirow}
\usepackage{subcaption}
\usepackage[numbers,sort&compress]{natbib}

\title{CARD: Correlation Aware Restoration with Diffusion}

\author{Niki Nezakati, Arnab Ghosh, Amit Roy-Chowdhury, Vishwanath Saragadam\\
University of California, Riverside \\
{\tt\small \{nneza001, aghos034, amitrc, vishwans\}@ucr.edu}
}

\usepackage{xcolor}

\begin{document}
\maketitle
\begin{abstract}
Denoising diffusion models have achieved state-of-the-art performance in image restoration by modeling the process as sequential denoising steps. However, most approaches assume independent and identically distributed (i.i.d.) Gaussian noise, while real-world sensors often exhibit spatially correlated noise due to readout mechanisms, limiting their practical effectiveness.
We introduce \textbf{C}orrelation \textbf{A}ware \textbf{R}estoration with \textbf{D}iffusion (CARD), a training-free extension of DDRM that explicitly handles correlated Gaussian noise. CARD first whitens the noisy observation, which converts the noise into an i.i.d. form. Then, the diffusion restoration steps are replaced with noise-whitened updates, which inherits DDRM's closed-form sampling efficiency while now being able to handle correlated noise.
To emphasize the importance of addressing correlated noise, we contribute CIN-D, a novel correlated noise dataset captured across diverse illumination conditions to evaluate restoration methods on real rolling-shutter sensor noise. This dataset fills a critical gap in the literature for experimental evaluation with real-world correlated noise. Experiments on standard benchmarks with synthetic correlated noise and on CIN-D demonstrate that CARD consistently outperforms existing methods across denoising, deblurring, and super-resolution tasks.
\end{abstract}
    
\section{Introduction}
\label{sec:intro}

\begin{figure}[!tt]
  \centering
\includegraphics[width=\columnwidth]{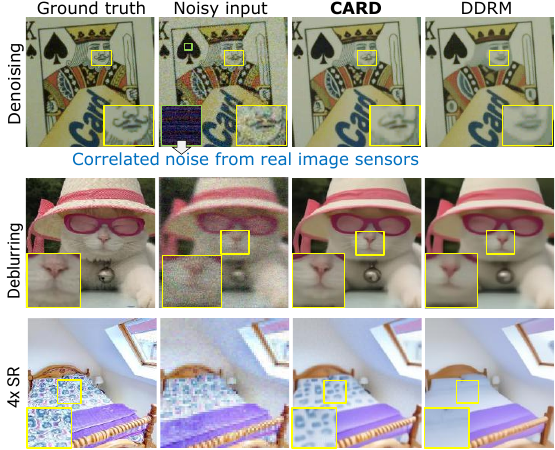}
  \caption{\textbf{Play your CARDs right.} Image sensors, particularly rolling-shutter ones, suffer from correlated noise. We propose CARD, a \textbf{training-free} diffusion-based method for solving inverse problems with correlated noise. CARD requires a covariance matrix estimate and a pre-trained diffusion model, and can solve linear inverse problems such as (top row) denoising, (middle row) deblurring, and (bottom row) super-resolution.}
  \label{fig:teaser}
\end{figure}

\begin{figure}[!tt]
  \centering
\includegraphics[width=\columnwidth]{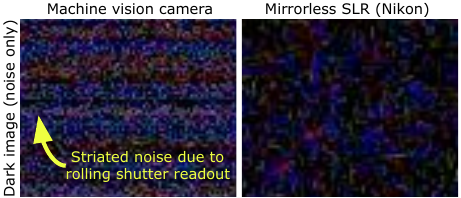}
  \caption{\textbf{Correlated noise in real cameras.} Modern digital cameras, particularly the rolling shutter ones, have strongly correlated noise. Here we show dark images for a machine vision, and a mirrorless SLR camera, both equipped with rolling shutter CMOS sensors, and displaying strong spatial correlations in noise. This observation motivates CARD and its applications for solving inverse problems with correlated noise.}
  \label{fig:intro-figure-noise}
\end{figure}

Image restoration is a fundamental problem in computer vision, with applications ranging from denoising and super-resolution to deblurring. 
A key assumption across most restoration approaches is that the added noise is independent and identically distributed (i.i.d.).
This assumption has been at the core of nearly all restoration approaches, whether classical~\cite{chambolle1997image,elad2006sparse,dabov2007bm3d}, feed forward neural network-based~\cite{mao2016skip,liang2021swinir,zamir2022restormer},  even self-supervised~\cite{ulyanov2018deep,wang2023noise2info}, and the more recent plug-and-play diffusion model-based techniques~\cite{kawar2022ddrm,wang2022ddnm}. 
However, this assumption is often violated on several modern digital cameras, particularly equipped with rolling shutter sensors, where data is read out in a row-wise manner.
Most cellphone cameras, and several single lens reflex (SLR) cameras are equipped with such rolling shutter sensors, implying that the added noise is no longer i.i.d., but often spatially, and sometimes temporally, correlated. This affects the performance of various image restoration and enhancement modules that are in-built into such cameras (Fig.~\ref{fig:teaser}). As shown in Fig.~\ref{fig:intro-figure-noise}, real rolling-shutter cameras exhibit strong inter-pixel correlations, visible both in dark frames and their estimated covariance matrices.
\textit{This mismatch between the i.i.d. assumption and the reality of consumer imaging systems represents a fundamental limitation for existing restoration methods.}

In this paper, we introduce \textit{Correlation Aware Restoration with Diffusion (CARD)}, a training-free framework for image restoration corrupted by spatially correlated noise.
CARD requires two key components, an estimate of the covariance matrix, and a pre-trained diffusion model.
CARD operates in two steps. First, the input image corrupted by correlated noise is \textit{whitened}.
This transformation converts the correlated observations into an equivalent independent and identically distributed (i.i.d.) form.
Second, \textit{CARD leverages a modified form of DDRM}, which enjoys closed-form conditioning steps for efficient computation.
Specifically, the standard diffusion updates are replaced by \emph{noise-whitened updates} using the covariance matrix.
The CARD approach is model-agnostic and can be readily integrated with any pretrained diffusion model used for linear inverse problems, including super-resolution, deblurring, and other linear restoration tasks.
We demonstrate across a variety of datasets and noise levels that CARD consistently outperforms prior denoising approaches that were trained primarily for i.i.d. or correlated noise.

To further validate CARD's efficacy in real settings, we have collected a new dataset, named Correlated Image Noise Dataset (CIN-D).
Several existing denoising datasets~\cite{abdelhamed2018sidd, xu2018polyu, plotz2017dnd} contain images with uncorrelated (i.i.d.) noise. 
To our knowledge, there is currently no publicly available dataset that explicitly models or contains spatially correlated noise in images.
To bridge this critical gap, we collected images across diverse illumination conditions (indoor and outdoor) to showcase the effects of spatially correlated noise.
CIN-D contains 400 images across 100 distinct scenes, with each scene containing three noise levels, and a noise-free image that acts as ground truth, and dark images for covariance matrix estimation.
CIN-D will serve as an evaluation bed for this paper, and future approaches addressing correlated noise. The complete CIN-D dataset and CARD implementation will be released to the public.
\section{Prior Work}
\label{sec:prior_work}
CARD draws inspiration from several classical approaches on denoising signals with correlated noise, and recent advances in diffusion-based restoration. Here, we provide a brief overview of the relevant literature.
\paragraph{Model-based image restoration.}
Classical methods treat restoration as regularized inverse problems using handcrafted priors. Total variation~\cite{rudin1992nonlinear,chambolle1997image} encourages piecewise smoothness, wavelet-based methods~\cite{portilla2003wavelet} promote sparsity, and non-local approaches~\cite{buades2005nonlocal,dabov2007bm3d} exploit patch recurrence. Dictionary learning~\cite{elad2006sparse} adapts sparse representations to the data. While effective for i.i.d. Gaussian noise, these methods struggle with spatially correlated sensor noise.

\paragraph{Learning-based image restoration.}
Deep networks have significantly improved restoration quality through supervised learning~\cite{zhang2017dncnn,ledig2017photo,haris2018dbpn,kupyn2019deblurgan}, attention mechanisms~\cite{suin2020spatially}, and plug-and-play frameworks~\cite{zhang2021dpir}. Self-supervised alternatives include Deep Image Prior~\cite{ulyanov2018deep}, which optimizes on individual images, and generative priors~\cite{bora2017compressed,menon2020pulse,pan2020deep} that restore through latent code optimization. Despite strong performance, most methods assume independent noise or require task-specific training. In contrast, our approach works with correlated noise and for any linear inverse problem.

\paragraph{Diffusion priors for restoration.}
Diffusion models~\cite{ho2020denoising,song2020score} provide powerful generative priors for restoration. Methods like ILVR~\cite{choi2021ilvr} inject low-frequency guidance, while DDRM~\cite{kawar2022ddrm} derives closed-form updates in the degradation operator's spectral domain. DDNM~\cite{wang2022ddnm} maintains consistency through null-space projections, DPS~\cite{chung2023dps} uses gradient-based guidance, and DiffIR~\cite{xia2023diffir} improves efficiency through compact representations. Our approach is closely related to DDRM, but goes beyond the i.i.d. noise assumption that is central to its working.

\paragraph{Correlated noise and real sensors.}
Real sensor noise exhibits spatial correlation due to optical effects, readout circuitry, and image signal processing~\cite{boyat2015noisereview,verma2013photonnoise,plotz2017dnd,lukas2006readoutnoise,chatterjee2011noise,jin2020demosaicking}. Physics-based models~\cite{wei2020physicscvpr,wei2021physics} simulate sensor statistics during training. Self-supervised methods~\cite{lehtinen2018noise2noise,krull2019noise2void,huang2021neighbor2neighbor} learn from noisy observations but typically assume independence. Recent work addresses correlation through information-theoretic supervision~\cite{wang2023noise2info}, spatially adaptive losses~\cite{li2023spatially}, burst-based pseudo-targets~\cite{vaksman2023pcst}, and adaptive residual denoising~\cite{kim2025apr}. These approaches require retraining and approximate correlations heuristically rather than modeling them directly. CARD extends DDRM to handle correlated noise without retraining, working as a drop-in replacement for linear restoration tasks.
\section{Challenges with Correlated Noise}
We first start by covering diffusion-based restoration and then talk about how correlations in noise pose a challenge to effectively solving inverse problems.

\subsection{Denoising Diffusion Restoration Model}
\label{sec:method-ddrm}
Denoising Diffusion Restoration Model (DDRM)~\cite{kawar2022ddrm} combines a pretrained denoising diffusion model~\cite{ho2020denoising} with a linear forward operator. The measurement model is
\begin{equation}
\mathbf{y} = H \mathbf{x}_{0} + \mathbf{z},\qquad \mathbf{z} \sim \mathcal{N}(0,\sigma_{y}^{2} I),
\end{equation}
where $\mathbf{y}\in\mathbb{R}^{m}$ is the degraded observation, $H\in\mathbb{R}^{m\times d}$ encodes the degradation process, $\mathbf{x}_{0}\in\mathbb{R}^{d}$ is the unknown clean image, and $\mathbf{z}$ is i.i.d. Gaussian noise with variance $\sigma_{y}^{2}$. A pretrained diffusion model provides a prior over natural images by gradually perturbing a clean image $\mathbf{x}_{0}$ into a sequence of increasingly noisy variables $\{\mathbf{x}_{t}\}_{t=0}^{T}$ according to a variance schedule,
\begin{equation}
q(\mathbf{x}_t \mid \mathbf{x}_0) = \mathcal{N}\!\big(\mathbf{x}_0,\sigma_t^2 I\big),
\end{equation}
with \(0=\sigma_0<\sigma_1<\cdots<\sigma_T\). A learned denoiser $f_{\theta}(\mathbf{x}_{t},t)$ approximates the reverse process that removes noise to reconstruct clean samples. DDRM performs restoration by conditioning this pretrained prior on the degraded observation $\mathbf{y}$ through closed-form posterior updates at each diffusion timestep. The conditioning is performed in the spectral basis of the degradation operator,
\begin{equation}
H = U S V^{\top},
\end{equation}
where $U$ and $V$ are orthonormal and $S$ is diagonal with singular values $s_{i}$. Variables are projected into this basis as
\begin{equation}
\bar{\mathbf{x}}_{t}=V^{\top} \mathbf{x}_{t},\qquad \bar{\mathbf{y}}=S^{\dagger} U^{\top} \mathbf{y},
\end{equation}
where $S^{\dagger}$ is the Moore--Penrose pseudoinverse. Under independent measurement noise, each spectral component evolves as a one-dimensional Gaussian, allowing DDRM to derive a closed-form update at each step $t$ that interpolates between the current clean estimate and the measurement. The interpolation weights depend on $\sigma_{t}$, $\sigma_{y}$, and the singular values $s_{i}$. During sampling, the unknown clean component is replaced by the network prediction 
\begin{equation}
\mathbf{x}_{\theta,t}=f_{\theta}(\mathbf{x}_{t+1},t+1), 
\end{equation}
enabling efficient posterior sampling without retraining, provided the measurement noise is i.i.d.

\subsection{Measurements with correlated noise}
\label{sec:method-obs_model}
Real sensors often produce correlated noise, violating the independence assumption in standard DDRM. To enable DDRM to handle such conditions, we extend its observation model to incorporate correlated noise directly. The measurement process becomes
\begin{equation}
\mathbf{y} = H \mathbf{x}_{0} + \mathbf{n},\qquad n \sim \mathcal{N}(0,\sigma_{y}^{2}\Sigma),
\end{equation}
where $\mathbf{n}\in\mathbb{R}^{m}$ is zero-mean Gaussian noise with covariance $\sigma_{y}^{2}\Sigma\in\mathbb{R}^{m\times m}$. The matrix $\Sigma$ is symmetric positive definite and captures spatial correlations which couple the spectral components and prevent the closed-form conditioning that DDRM relies on. 
\section{The CARD Method}
\label{sec:method}

\begin{figure}[!tt]
  \centering
  \includegraphics[width=1.0\linewidth]{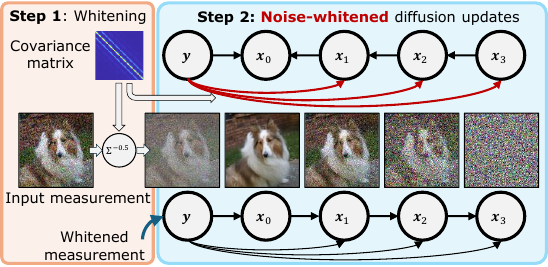}
  \caption{\textbf{CARD approach.} CARD is a two-step training-free approach to solving image-based inverse problems corrupted by correlated noise, as often found in real cameras. In step 1, we whiten the measurement using a known covariance matrix. In step 2, we apply noise-whitened updates to a pre-trained unconditional diffusion model to solve the whitened inverse problem.}
  \label{fig:method-figure}
\end{figure}

Our approach, correlation-aware restoration with diffusion (CARD) is a two step process illustrated in Fig.~\ref{fig:method-figure}. First we whiten the image to restore the i.i.d. form of the noise. Second, we use a modified diffusion-based restoration. Specifically, we modify the data-consistency step by preconditioning the measurements to ensure the noise independence is satisfied. CARD hence requires an estimate of the covariance matrix $\Sigma$ and a pre-trained diffusion model without any task-specific training. 

\paragraph{CARD step 1: Whitening.}
\label{sec:method-whitening}
We restore independence in noise by preconditioning the measurement equation. The whitened model is obtained by left-multiplying with $\Sigma^{-1/2}$:
\begin{equation}
\Sigma^{-1/2} \mathbf{y} \;=\; \Sigma^{-1/2} H \mathbf{x}_{0} \;+\; \Sigma^{-1/2} \mathbf{n},
\end{equation}
where $\Sigma^{-1/2}$ is the symmetric inverse square root of $\Sigma$. The preconditioned variables become
\begin{equation}
\tilde{\mathbf{y}}=\Sigma^{-1/2} \mathbf{y},\qquad \tilde{H}=\Sigma^{-1/2} H,\qquad \tilde{\mathbf{n}}=\Sigma^{-1/2} \mathbf{n},
\end{equation}
where $\tilde{n}$ is now i.i.d. Gaussian. The resulting measurement,
\begin{equation}
\tilde{\mathbf{y}} = \tilde{H} \mathbf{x}_{0} + \tilde{n},\qquad \tilde{\mathbf{n}}\sim\mathcal{N}(0,\sigma_{y}^{2}I),
\end{equation}
satisfies DDRM's independence requirement. In practice, we compute $\Sigma^{-1/2}$ efficiently using the Cholesky factorization $LL^{\top}=\Sigma$, forming $W=L^{-1}$ such that $W\Sigma W^{\top}=I$.

\paragraph{CARD step 2: Noise-whitened updates.}With a modified measurement matrix and i.i.d. noise, we apply DDRM in this whitened measurement space by computing the singular value decomposition of the whitened operator,
\begin{equation}
\tilde{H} = \tilde{U} \tilde{S} \tilde{V}^{\top},
\end{equation}
where $\tilde{U}$ and $\tilde{V}$ are orthonormal and $\tilde{S}$ is diagonal with singular values $\tilde{s}_{i}$. The spectral coordinates are
\begin{equation}
\bar{\tilde{\mathbf{x}}}_{t}=\tilde{V}^{\top} \mathbf{x}_{t},\qquad \bar{\tilde{\mathbf{y}}}=\tilde{S}^{\dagger} \tilde{U}^{\top} \tilde{\mathbf{y}}.
\end{equation}
In DDRM, the time-step update depends explicitly on the singular values $s_{i}$ of the forward operator $H$. Instead, we apply DDRM in the whitened measurement space using the singular values $\tilde{s}_{i}$ of $\tilde{H}$. Full whitened update equations are provided in the supplementary material.

\paragraph{Practical considerations.} For high-resolution images, we apply whitening and noise sampling patchwise over non-overlapping tiles. Sensor noise correlations are typically short-range, decaying beyond a few neighboring pixels, which makes patchwise processing a reasonable approximation. We discuss the effect of patch size in the supplementary material of this paper.
\section{Correlated Image Noise Dataset (CIN-D)}
\label{sec:dataset}

\begin{figure}[!tt]
  \centering
  \includegraphics[width=1.0\linewidth]{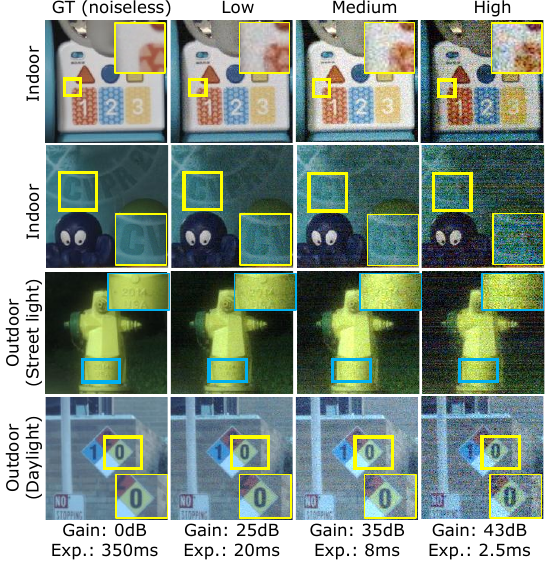}
  \caption{\textbf{Example images from CIN-D dataset.} This paper contributes a new dataset with varying levels of correlated noise, captured with a FLIR Blackfly machine vision camera. The dataset contains curated indoor and outdoor scenes for evaluating restoration methods under realistic correlated noise.}
  
  \label{fig:datasetCornoise_BFS100}
\end{figure}

Existing denoising datasets~\cite{abdelhamed2018sidd, plotz2017dnd} typically assume i.i.d. sensor noise, and thereby nearly all denoising approaches focus primarily on removing uncorrelated noise.
However, real sensors often exhibit spatially correlated noise~\cite{ringaby2014geometric}, particularly in rolling-shutter systems (Fig.~\ref{fig:intro-figure-noise}).
To address this limitation, we introduce CIN-D (Correlated Image Noise Dataset).
CIN-D contains clean and noisy images captured with a rolling-shutter camera, providing realistic correlated noise patterns that demonstrate CARD's practical value.

\paragraph{Acquisition} CIN-D is captured using a FLIR Blackfly S BFS-U3-63S4C-C color camera, comprising 100 scenes at three noise levels (low, medium, high), and an additional long exposure shot that acts as noiseless ground truth. We jointly vary gain and exposure while adjusting the lens f-number to vary the noise strength (Fig.\ref{fig:datasetCornoise_BFS100}). CIN-D scenes include fluorescent-lit indoor and outdoor daylight/nighttime environments, all static to avoid motion. At higher gains, the captured images exhibits structured artifacts such as row banding that challenge established restoration methods, and motivates the need for the CARD approach. Further details about the CIN-D are available in the supplementary.

\paragraph{Estimation of noise covariance matrix.}
We estimate the spatially correlated noise of the camera sensor from dark frames captured at the same exposure and gain settings as the scene images, with the lens cap on. We model the dark-frame signal as a zero-mean random process that is approximately locally stationary over small patches. We first normalize intensities to $[0,1]$, and convert to a single-channel linear luminance image $L$ to obtain a scalar representation of the noise, $L = 0.2126\,R + 0.7152\,G + 0.0722\,B$.
We partition the luminance image into non-overlapping $8\times8$ patches and vectorize each patch as $\mathbf{z}\in\mathbb{R}^{64}$. We aggregate  all patches from all dark-frames to yield \(N\) samples, which we stack to form the data matrix
$ Z = [\mathbf{z}_1,\dots,\mathbf{z}_N] \in \mathbb{R}^{64\times N}.
$
After we subtract the per-pixel mean from each row of \(Z\), we estimate the sample covariance $\Sigma\in \mathbb{R}^{64\times 64}$. The noise covariance matrix  $\Sigma$ captures spatial correlations between pixels within local neighborhoods. We use this estimated covariance for the noise whitening step in CARD when evaluating on real data (Section~\ref{sec:method}).

\section{Experiments}
\label{sec:experiments}
We evaluate CARD on three linear restoration tasks, including denoising under correlated Gaussian noise, deblurring with uniform, Gaussian, and anisotropic kernels, and $2\times$ and $4\times$ super-resolution. We report peak signal-to-noise ratio (PSNR) and LPIPS for perceptual quality. SSIM results are provided in the supplementary material of this paper. 

\subsection{Noise modeling and whitening}
We evaluate on ImageNet~\cite{deng2009imagenet}, LSUN-Bed, LSUN-Cat ~\cite{yu2015lsun} validation images, and our captured dataset (CIN-D), all at $256\times256$ resolution and preprocessed following DDRM to ensure consistency. For evaluation on standard benchmarks, we inject synthetic correlated Gaussian noise $\mathbf{n} \sim \mathcal{N}(0, \Sigma_{\text{synth}})$, where,
\begin{equation}
\Sigma_\text{synth} = \sigma^{2}(I + \alpha B) + \varepsilon I,
\end{equation}
where $B$ is a symmetric banded matrix with non-zeros on selected off-diagonals, $\alpha$ controls the correlation strength, and $\varepsilon$ is a small regularization term ensuring positive definiteness. 
This covariance matrix closely resembles $\Sigma$ estimated from real sensor noise with correlations among adjacent pixels.
Given $\Sigma_\text{synth}$, correlated samples are generated as $\mathbf{n}=L\mathbf{z}$ with $\mathbf{z}\!\sim\!\mathcal{N}(0,I)$ and $L$ the Cholesky factor of $\Sigma_\text{synth}$. 
We then perform inference in the whitened measurement domain using $W = \Sigma_{\text{synth}}^{-1/2}$. We use simulated correlated noise to evaluate CARD on standard benchmarks such as ImageNet and LSUN, enabling direct comparison with prior work. To evaluate performance on real data, we test CARD on CIN-D using the estimated covariance $\Sigma$ described in section~\ref{sec:dataset}.

\subsection{Degradation models}
For denoising, the forward operator is the identity matrix ($H=I$). For deblurring, we consider three point-spread functions, including a $9\times9$ uniform kernel, an isotropic Gaussian kernel, and an anisotropic Gaussian kernel. Very small singular values of the blur operators are set to zero to mitigate ill-posedness. For super-resolution, we use block-averaging downsamplers with scale factors of $2\times$ and $4\times$, applied independently along each axis. For experiments on ImageNet and LSUN, we add synthetic zero-mean correlated Gaussian noise with covariance $\Sigma_\text{synth}$ to corrupted measurements. 

\subsection{Implementation details}
We build on the public DDRM codebase, incorporating the whitening step and performing all updates in the whitened measurement space. All hyperparameters match those of DDRM except for the measurement-blend parameter $\eta_b$, which we optimized for best performance on images with correlated noise. Based on a grid search over a held-out validation split, we set $\eta=0.80$ and $\eta_b=1.0$. We use the same uniformly spaced timestep schedule as the 1000-step pretrained models and perform 20 neural function evaluations (NFEs) per sample. For simulated experiments, we use a consistent noise level $\sigma_0$ across all methods. For real-world experiments, we tune $\sigma_0$ separately for each baseline to ensure fair comparisons. Further implementation details are added in the supplementary material of this paper.

\begin{table}[!tt]
\caption{\textbf{Denoising results on ImageNet and LSUN} across low (0.1), medium (0.5), and high (0.9) noise levels. We report PSNR/LPIPS as P/L. Best and second‐best results are marked in bold and underlined, respectively. CARD achieves the best performance across both datasets and all noise levels.}
\label{tab:deno-imagenet-lsun}
\centering

\begin{subtable}[!tt]{\columnwidth}
\centering
\caption{ImageNet}
\resizebox{\linewidth}{!}{
\begin{tabular}{@{}llccc@{}}
\toprule
\multirow{2}{*}{Category} & \multirow{2}{*}{Model} & \multicolumn{3}{c}{Noise Level $\sigma_0$} \\
 & & 0.1 (P/L) & 0.5 (P/L) & 0.9 (P/L) \\
\midrule
\multirow{4}{*}{\parbox[t]{2.8cm}{\raggedright Learning-Based\\(i.i.d.)}}
& Restormer~\cite{zamir2022restormer}   & 10.5/0.78 & 10.3/0.96 & 10.1/1.14 \\
& DnCNN~\cite{zhang2017dncnn}         & 22.6/0.24 & 20.4/0.46 & 18.1/0.66 \\
& Noise2Info~\cite{wang2023noise2info}& 25.2/0.24 & 24.6/\underline{0.26} & \underline{23.3}/\underline{0.29} \\
& PCST~\cite{vaksman2023pcst}         & 25.6/0.36 & 19.8/0.63 & 16.6/0.79 \\
\midrule
\multirow{3}{*}{\parbox[t]{2.8cm}{\raggedright Prior-Based\\(i.i.d.)}}
& BM3D~\cite{dabov2007bm3d}           & 30.1/\underline{0.11} & \underline{25.8}/0.33 & 22.4/0.50 \\
& DDNM~\cite{wang2022ddnm}            & 28.1/0.25 & 16.3/0.57 & 12.3/0.68 \\
& DDRM~\cite{kawar2022ddrm}         & \underline{31.0}/0.14 & 24.8/0.33 & 22.7/0.40 \\
\midrule
\multirow{2}{*}{\parbox[t]{2.8cm}{\raggedright Learning-Based\\(correlated)}}
& APRRD-BSN~\cite{kim2025apr}        & 23.9/0.25 & 22.0/0.38 & 20.1/0.51 \\
& APRRD-NBSN~\cite{kim2025apr}       & 24.9/0.31 & 22.4/0.45 & 20.5/0.54 \\
\midrule
\multirow{1}{*}{\parbox[t]{2.8cm}{\raggedright Ours}}
& \textbf{CARD}                       & \textbf{34.0}/\textbf{0.07} & \textbf{29.1}/\textbf{0.15} & \textbf{26.7}/\textbf{0.22} \\
\bottomrule
\end{tabular}
}
\end{subtable}
\hfill

\vspace{0.2em}
\begin{subtable}[!tt]{\columnwidth}
\centering
\caption{LSUN (Bed + Cat Average)}
\resizebox{\linewidth}{!}{
\begin{tabular}{@{}llccc@{}}
\toprule
\multirow{2}{*}{Category} & \multirow{2}{*}{Model} & \multicolumn{3}{c}{Noise Level $\sigma_0$} \\
 & & 0.1 (P/L) & 0.5 (P/L) & 0.9 (P/L) \\
\midrule
\multirow{4}{*}{\parbox[t]{2.4cm}{\raggedright Learning-Based\\(i.i.d.)}}
& Restormer~\cite{zamir2022restormer} & 10.3/0.78 & 10.2/0.97 & 10.1/1.15 \\
& DnCNN~\cite{zhang2017dncnn}         & 23.6/0.20 & 21.2/0.45 & 18.6/0.67 \\
& Noise2Info~\cite{wang2023noise2info}& 25.9/0.23 & 25.4/\underline{0.24} & 23.9/\underline{0.29} \\
& PCST~\cite{vaksman2023pcst}         & 26.2/0.34 & 20.2/0.63 & 16.9/0.79 \\
\midrule
\multirow{3}{*}{\parbox[t]{2.4cm}{\raggedright Prior-Based\\(i.i.d.)}}
& BM3D~\cite{dabov2007bm3d}           & 30.1/0.14 & 26.1/0.32 & 22.4/0.50 \\
& DDNM~\cite{wang2022ddnm}            & 28.4/0.25 & 16.4/0.56 & 12.3/0.67 \\
& DDRM~\cite{kawar2022ddrm}           & \underline{31.9/0.12} & \underline{26.0}/0.27 & \underline{24.0}/0.32 \\
\midrule
\multirow{2}{*}{\parbox[t]{2.4cm}{\raggedright Learning-Based\\(correlated)}}
& APRRD-BSN~\cite{kim2025apr}         & 24.3/0.26 & 22.6/0.37 & 20.8/0.50 \\
& APRRD-NBSN~\cite{kim2025apr}        & 25.6/0.31 & 23.1/0.45 & 21.2/0.55 \\
\midrule
\multirow{1}{*}{\parbox[t]{2.4cm}{\raggedright Ours}}
& \textbf{CARD}                       & \textbf{34.7}/\textbf{0.06} & \textbf{29.6}/\textbf{0.14} & \textbf{27.2}/\textbf{0.19} \\
\bottomrule
\end{tabular}
}
\end{subtable}

\end{table}

\begin{figure*}[!tt]
  \centering
  \includegraphics[width=1.0\linewidth]{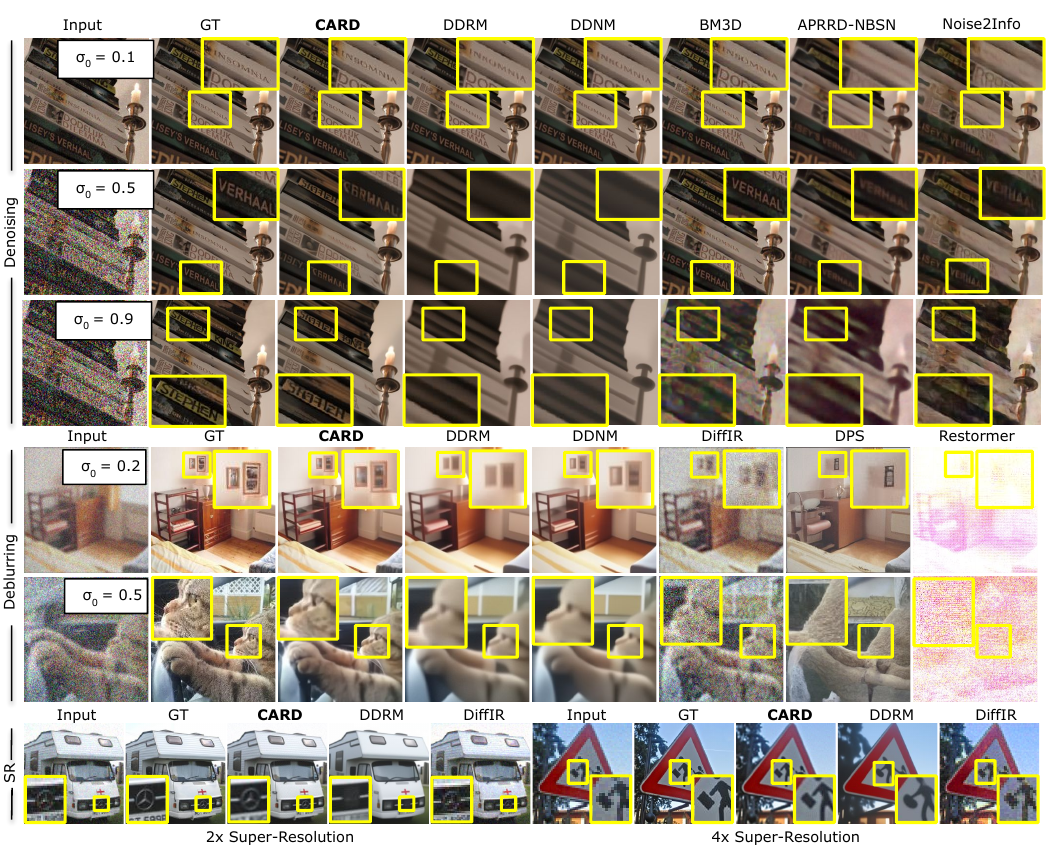}
  \caption{\textbf{Qualitative comparison of denoising, gaussian deblurring, and super-resolution.} Results are shown for denoising at varying noise levels ($\sigma_0=0.1, 0.5, 0.9$), gaussian deblurring at $\sigma_0=0.2 , 0.5$, and super-resolution ($2\times$ and $4\times$) at $\sigma_0=0.2$. The boxed regions highlight interesting areas. CARD removes noise while preserving fine details, outperforming other baselines.}
  \label{fig:exp-deno-figure}
\end{figure*}

\subsection{Experiments with simulated correlated noise}
\paragraph{Denoising results.}
Tables~\ref{tab:deno-imagenet-lsun}(a) and~\ref{tab:deno-imagenet-lsun}(b) present denoising results on ImageNet and LSUN (averaged over LSUN-Bed and LSUN-Cat) at three noise levels ($\sigma_0{=}0.1$, $0.5$, $0.9$). CARD  outperforms all baselines across datasets and noise levels. At low noise ($\sigma_0{=}0.1$), CARD surpasses the strongest i.i.d. diffusion baseline DDRM. The advantage increases at medium noise ($\sigma_0{=}0.5$), where learning-based methods deteriorate and DDRM becomes less stable. At high noise ($\sigma_0{=}0.9$), CARD maintains high PSNR and low LPIPS while existing methods fail, demonstrating strong robustness to severe correlated noise. Fig.~\ref{fig:exp-deno-figure} (Denoising) shows a representative ImageNet example where, while most methods perform well at low noise, CARD preserves fine details such as text on books as noise increases, whereas competing methods progressively lose structure.

\begin{table}[!tt]
\caption{\textbf{Gaussian deblurring results on ImageNet and LSUN} across medium (0.2) and high (0.5) noise levels. We report PSNR/LPIPS as P/L. Best and second‐best results are marked in bold and underlined, respectively. CARD achieves the best performance on both datasets and at both noise levels.}
\label{tab:deblur-gaussian-all}
\centering
\resizebox{\columnwidth}{!}{
\begin{tabular}{@{}lcccc@{}}
\toprule
\multirow{2}{*}{Model} &
\multicolumn{2}{c}{ImageNet} &
\multicolumn{2}{c}{LSUN (Bed + Cat Average)} \\
& 0.2 (P/L) & 0.5 (P/L) & 0.2 (P/L) & 0.5 (P/L) \\
\midrule
Restormer        
& 7.8/0.94  & 8.0/1.08  & 8.1/0.92 & 8.3/1.07 \\

DPS              
& 19.4/0.41 & 19.3/0.46 & 19.4/0.40 & 19.1/0.45 \\

DiffIR           
& 20.1/0.65 & 17.8/0.98 & 20.1/0.70 & 17.9/1.03 \\

DDNM             
& 21.8/0.45 & 15.9/0.60 & 22.1/0.43 & 16.0/0.59 \\

DDRM             
& \underline{24.3/0.36} & \underline{22.2/0.42}
& \underline{25.3/0.30} & \underline{23.1/0.36} \\

\textbf{CARD (ours)} 
& \textbf{26.6}/\textbf{0.23} & \textbf{24.4}/\textbf{0.30}
& \textbf{27.1}/\textbf{0.20} & \textbf{24.8}/\textbf{0.31} \\
\bottomrule
\end{tabular}
}
\end{table}

\paragraph{Deblurring results.}
Table~\ref{tab:deblur-gaussian-all} presents Gaussian deblurring results on ImageNet and LSUN (averaged over LSUN-Bed and LSUN-Cat) at two noise levels ($\sigma_0{=}0.2$ and $0.5$). CARD consistently outperforms all baselines, achieving PSNR improvements over the next-best method, DDRM, while maintaining lower LPIPS scores. Learning-based methods like Restormer struggle with stronger noise, and diffusion posteriors (DPS, DDNM) show limited robustness as noise increases. CARD maintains stable and superior performance across all conditions. Fig.~\ref{fig:exp-deno-figure} (Deblurring) shows qualitative comparisons at both noise levels. CARD restores fine details such as object structures and cat features with sharper edges and clearer textures. Baselines either over-smooth or leave residual blur, especially at $\sigma_0{=}0.5$, while CARD maintains consistent quality.

\begin{table}[!tt]
\caption{\textbf{Super-resolution results on ImageNet and LSUN} across $2\times$ and $4\times$ upscaling, medium (0.2) and high (0.5) noise levels. We report PSNR/LPIPS as P/L. Best and second-best results are marked in bold and underlined, respectively. CARD achieves the best performance on both datasets, for both $\times$2 and $\times$4 upscaling, and at both noise levels.}

\label{tab:sr-imagenet-lsun}
\centering

\begin{subtable}[!tt]{\columnwidth}
\centering
\caption{ImageNet}
\resizebox{\linewidth}{!}{
\begin{tabular}{@{}lcccc@{}}
\toprule
\multirow{2}{*}{Model} & \multicolumn{2}{c}{SR2} & \multicolumn{2}{c}{SR4} \\
& 0.2 (P/L) & 0.5 (P/L) & 0.2 (P/L) & 0.5 (P/L) \\
\midrule

DiffIR        
& 25.7/0.34 & 23.9/0.42 & 23.8/0.47 & 23.0/0.60 \\

DDRM          
& \underline{25.7/0.32} & \underline{22.7/0.41} 
& \underline{22.9/0.40} & \underline{20.4/0.46} \\

\textbf{CARD (ours)} 
& \textbf{28.0}/\textbf{0.20} 
& \textbf{25.1}/\textbf{0.33} 
& \textbf{25.0}/\textbf{0.30} 
& \textbf{23.1}/\textbf{0.40} \\

\bottomrule
\end{tabular}
}
\end{subtable}

\vspace{0.75em} 

\begin{subtable}[!tt]{\columnwidth}
\centering
\caption{LSUN (Bed + Cat Average)}
\resizebox{\linewidth}{!}{
\begin{tabular}{@{}lcccc@{}}
\toprule
\multirow{2}{*}{Model} &
\multicolumn{2}{c}{SR2} &
\multicolumn{2}{c}{SR4} \\
& 0.2 (P/L) & 0.5 (P/L) & 0.2 (P/L) & 0.5 (P/L) \\
\midrule

DiffIR
& 27.2/0.33 & 23.8/0.63 & 25.0/0.44 & 24.0/0.59 \\

DDRM
& \underline{26.3/0.27} & \underline{23.6/0.34} 
& \underline{23.4/0.33} & \underline{21.0/0.39} \\

\textbf{CARD (ours)}
& \textbf{29.0}/\textbf{0.16} 
& \textbf{26.9}/\textbf{0.24}
& \textbf{26.2}/\textbf{0.25}
& \textbf{25.1}/\textbf{0.31} \\

\bottomrule
\end{tabular}
}
\end{subtable}

\end{table}

\begin{table}[!tt]
\caption{\textbf{Denoising results on our dataset (CIN-D)}, evaluated at low and high noise levels. We report PSNR/LPIPS as P/L. Best and second‐best results are marked in bold and underlined, respectively. CARD achieves the best performance at both noise levels.}
\label{tab:deno-ourdata}
\centering
\resizebox{\columnwidth}{!}{
\begin{tabular}{@{}llcc@{}}
\toprule
\multirow{2}{*}{Category} & \multirow{2}{*}{Model} & \multicolumn{2}{c}{Noise Level} \\
 & & Low (P/L) & High (P/L) \\
\midrule
\multirow{4}{*}{\parbox[t]{2.8cm}{\raggedright Learning-Based\\(i.i.d.)}}
& Restormer~\cite{zamir2022restormer}   & 17.2/0.77 & 16.3/0.96 \\
& DnCNN~\cite{zhang2017dncnn}       & 28.6/0.29 & 24.7/0.67 \\
& Noise2Info~\cite{wang2023noise2info}  & 16.3/0.35 & 17.0/0.41 \\
& PCST~\cite{vaksman2023pcst}        & \underline{37.2}/\underline{0.29} & 29.4/\underline{0.38} \\
\midrule
\multirow{3}{*}{\parbox[t]{2.8cm}{\raggedright Prior-Based\\(i.i.d.)}}
& BM3D~\cite{dabov2007bm3d}        & 33.7/0.37 & 26.6/0.69 \\
& DDNM~\cite{wang2022ddnm}        & 33.9/0.33 & 24.8/0.50 \\
& DDRM~\cite{kawar2022ddrm}        & 35.3/\underline{0.29} & \underline{30.8}/0.39 \\
\midrule
\multirow{2}{*}{\parbox[t]{2.8cm}{\raggedright Learning-Based\\(correlated)}}
& APRRD-BSN~\cite{kim2025apr}   & 17.4/0.33 & 17.7/0.39 \\
& APRRD-NBSN~\cite{kim2025apr}  & 17.0/0.30 & 17.1/\underline{0.38} \\
\midrule
\multirow{1}{*}{\parbox[t]{2.8cm}{\raggedright Ours}}
& \textbf{CARD} & \textbf{38.1}/\textbf{0.28} & \textbf{31.5}/\textbf{0.34} \\
\bottomrule
\end{tabular}
}
\end{table}

\paragraph{Super-Resolution results.}
Tables~\ref{tab:sr-imagenet-lsun}(a) and~\ref{tab:sr-imagenet-lsun}(b) present super-resolution results on ImageNet and LSUN (averaged over LSUN-Bed and LSUN-Cat) for $2\times$ and $4\times$ upsampling at two noise levels ($\sigma_0{=}0.2$ and $0.5$). CARD consistently outperforms DiffIR and DDRM across all settings. For $2\times$ upsampling, CARD achieves 2-3 dB PSNR improvements over DDRM while maintaining lower LPIPS scores. The advantage increases for $4\times$ upsampling, demonstrating superior robustness to the more challenging task. At higher noise ($\sigma_0{=}0.5$), CARD maintains strong metrics while DiffIR and DDRM degrade substantially in both PSNR and LPIPS. Fig.~\ref{fig:exp-deno-figure} (SR) shows qualitative comparisons at $\sigma_0{=}0.2$. CARD reconstructs fine details such as car structures and traffic signs with sharper edges and clearer features. DiffIR and DDRM struggle to preserve these details, especially at $4\times$ upsampling.

\begin{figure*}[!tt]
  \centering
  \includegraphics[width=1.0\linewidth]{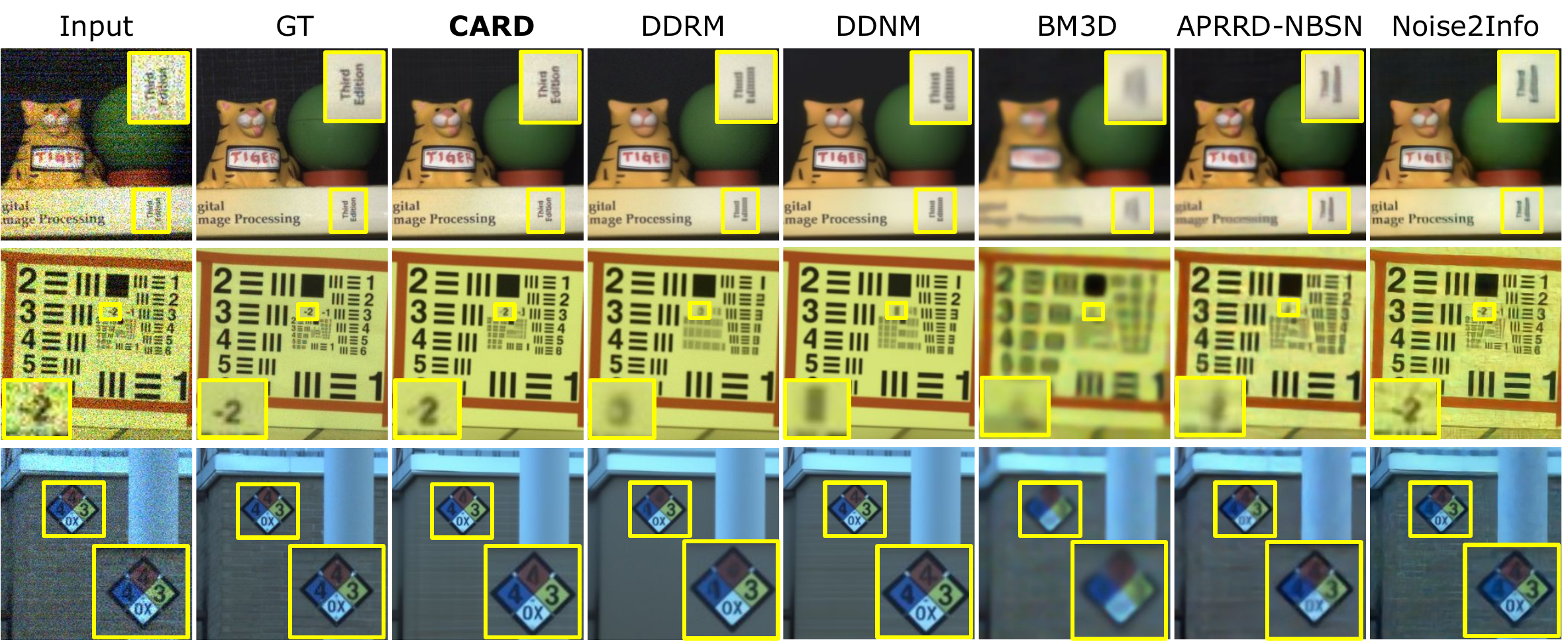}
    \caption{\textbf{Qualitative comparison of denoising results on our dataset (CIN-D).} The top two rows show indoor scenes, and the bottom row shows outdoor scene. Note that APRRD-NBSN and Noise2Info produce brighter results, contributing to their lower PSNR values. For visual clarity and fair comparison, we adjust the brightness of other methods to match. The boxed regions highlight interesting areas. CARD effectively suppresses noise while preserving fine details across both indoor and outdoor settings.}
  \label{fig:exp-deno-real-figure}
\end{figure*}

\subsection{Experiments on CIN-D}
\paragraph{Denoising results.}
Table \ref{tab:deno-ourdata} presents denoising results on our dataset (CIN-D). CARD consistently outperforms all baselines. It achieves PSNR gains over prior-based methods at both low and high noise levels while maintaining lower LPIPS scores. Learning-based methods trained on i.i.d. noise (PCST, DnCNN) perform well at low noise but degrade significantly at high noise levels. Correlated denoisers (APRRD-BSN, APRRD-NBSN) fail to generalize beyond their training statistics. Additional results for super-resolution and deblurring on CIN-D are provided in the supplementary material of this paper.

\section{Robustness of CARD}
\label{sec:ablation}

In this section, we analyze the robustness of CARD through two practical aspects. First, we examine sensitivity to errors in the estimated covariance matrix. Second, we investigate whether covariance estimates generalize across different sensors. Additional ablations on patch size for covariance estimation are provided in the supplementary material.

\begin{figure}[t]
  \centering
   \includegraphics[width=1.0\linewidth]{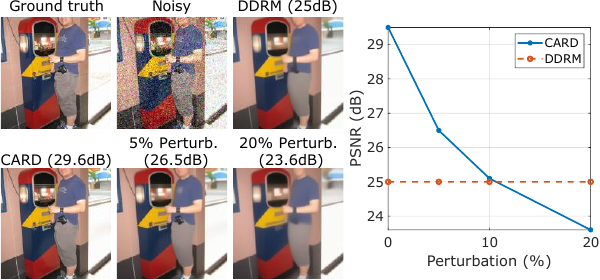}
   \caption{\textbf{Sensitivity of CARD to covariance matrix}. The ImageNet examples show qualitative results for GT, Noisy, DDRM, and CARD variants with increasing perturbation levels, with PSNR values indicated. The plot shows PSNR as a function of perturbation level. CARD remains robust and outperforms DDRM even with up to 10\% perturbation of the whitening matrix.}
   \label{fig:ablation-figure}
\end{figure}

\paragraph{Sensitivity to covariance matrix estimate.}
\label{sec:sensitivity-sigma}
We analyze the robustness of CARD to errors in the estimated noise covariance used for whitening.
To emulate imperfect calibration, we generate a synthetic noise with covariance matrix $\Sigma$ and then perturb the  whitening transformation by adding random symmetric noise to $\Sigma$ before computing its whitening transform.
The perturbation level, expressed as a percentage, controls the magnitude of injected noise relative to $\Sigma$. As shown in Fig.~\ref{fig:ablation-figure}, at perfect covariance knowledge (0\% perturbation), CARD achieves 29.5 dB compared to DDRM's 25.0 dB. Small perturbations (5\%) cause only minor PSNR drops to 26.5 and 25.1 dB, indicating stability under realistic estimation errors. At moderate to high perturbation (20\%), CARD degrades to 23.6 dB, yet maintains visually acceptable quality. This shows that CARD provides substantial improvements when covariance estimates are reasonable, while degrading gracefully as estimation quality decreases.

\begin{figure}
    \centering
    \includegraphics[width=\columnwidth]{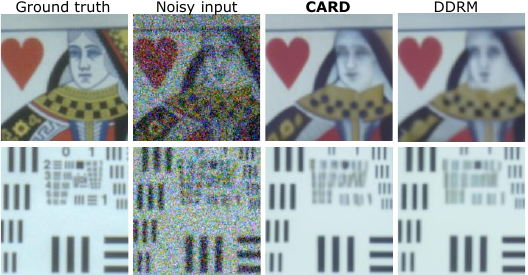}
    \caption{\textbf{Generalizability of CIN-D $\Sigma$ estimate.} In the figure above, we applied the covariance matrix estimated from the machine vision camera, to a Nikon Z30 mirrorless SLR camera. Even though $\Sigma$ was not estimated from Z30 images, CARD still produces better results than DDRM, showing that our estimate of $\Sigma$ from CIN-D generalizes well to other rolling shutter cameras.}
    \label{fig:z30}
\end{figure}

\paragraph{Generalization of CIN-D covariance estimate.}
We investigate whether the covariance matrix $\Sigma$ estimated from one sensor generalizes to other cameras. Specifically, we used the covariance matrix from the Blackfly machine vision dark frames (CIN-D) on a Nikon Z30 mirrorless camera. As shown in Fig.~\ref{fig:z30}, this domain transferred $\Sigma$ outperforms DDRM even though we do not explicitly use the covariance matrix from Z30. 
CARD likely depends more on the structure of $\Sigma$ (relative correlations) than its absolute magnitude. After scaling with the task noise level $\sigma_0$, even a modest but correctly-shaped covariance may be sufficient. Since $\Sigma_{\text{Blackfly}}$ preserves this structure, it transfers well to both CIN-D and Z30 data, demonstrating that the spatial correlation pattern of rolling-shutter noise is largely sensor-agnostic.

\section{Conclusion}
\label{sec:conclusion}
We introduced CARD, a training-free approach that adapts DDRM to explicitly handle spatially correlated Gaussian noise in image restoration. By whitening noisy observations and applying noise-whitened updates, CARD transforms correlated noise into an i.i.d. form while preserving DDRM's closed-form sampling efficiency. Experiments across denoising, deblurring, and super-resolution demonstrate that CARD consistently outperforms existing methods on both synthetic benchmarks and real sensor data. An important contribution of this work is CIN-D, a novel dataset captured with a rolling-shutter sensor across diverse illumination conditions that provides realistic correlated noise patterns for rigorous evaluation. While our experiments focused on RGB sensors, CARD is applicable to any imaging sensor exhibiting spatially correlated noise. Future work will explore estimating noise covariance directly from noisy images without requiring dark frames, which would enable broader deployment across different sensor types and camera models.

\noindent \textbf{Acknowledgment:}
This work is supported in part by NSF grants CNS-2312395 and CMMI-2326309, the SoCalHub initiative from the University of California Office of the President, and a UC Regents Faculty Fellowship.

{
    \small
    \bibliographystyle{unsrtnat}
    \bibliography{references}

}

\newpage
\setcounter{page}{1}

\newcommand{\beginsupplement}{%
        \setcounter{table}{0}
        \setcounter{figure}{0}
        \setcounter{section}{0}
        \setcounter{equation}{0}
     }

\beginsupplement

\twocolumn[
\begin{center}
    {\Large \bf Supplementary Material \\ }
    \vspace{5.0em}
\end{center}
]

\label{sec:supplementary}

\section{Implementation Details}
All experiments utilize frozen diffusion backbones identical to those in DDRM~\cite{kawar2022ddrm}. We employ publicly available UNet architectures at $256\times256$ resolution, the standard checkpoints for LSUN-Bedroom and LSUN-Cat, and the OpenAI/Guided-Diffusion models for ImageNet.

\subsection{Sampling configuration}
Following DDRM, we sample from a uniformly spaced subset of the $T=1000$ pretraining timesteps. We perform $K=20$ neural function evaluations (NFEs) per image through linear subsampling of the original schedule. Through hyperparameter grid search, we identify optimal values of $\eta=0.80$ for the stochasticity parameter and $\eta_b=1.0$ for the measurement-blending parameter. Compared to DDRM's default $\eta=0.85$, our reduced stochasticity consistently improves performance with correlated noise while preserving perceptual quality.

\subsection{Degradation models}
\label{supp:imp_deg_models}
We adopt the standard linear measurement model 
\begin{equation}
\mathbf{y}=H\mathbf{x}+\mathbf{n},
\end{equation}
with pixel values normalized to $[0,1]$. The forward operators $H$ are defined as follows. For denoising, we set $H=I$ (identity operator). For deblurring, we apply separable 1D kernels along horizontal and vertical axes with three blur types: (i) uniform box filter with length 9 (effective $9\times9$ PSF), (ii) isotropic Gaussian blur with 5-tap kernel ($\sigma=10$ in code units), and (iii) anisotropic Gaussian blur with orthogonal 9-tap kernels of differing spreads ($\sigma_x=20$, $\sigma_y=1$). To address ill-posedness, singular values below $3\times10^{-2}$ are set to zero following standard deblurring implementations. For super-resolution, we employ block-averaging downsamplers with scale factors $2\times$ and $4\times$, applied independently along each spatial dimension. All operators act on $[0,1]$-normalized images and remain fixed across all compared methods.

\subsection{Correlated noise generation and whitening}

We model synthetic correlated noise using the covariance structure
\begin{equation}
\Sigma_{\text{synth}} = \sigma^{2}(I + \alpha B) + \varepsilon I,  
\end{equation}
where $B$ is a symmetric banded adjacency matrix with non-zeros on selected off-diagonals. To maintain computational efficiency, correlation and whitening are applied to non-overlapping $8\times8$ patches per channel. For patch size $p=8$ and dimension $d=p^{2}=64$, we instantiate a $d\times d$ covariance matrix $\Sigma$ and compute its Cholesky factor $L$ and symmetric inverse square-root $W=\Sigma^{-1/2}$. Correlated noise samples are generated as $\mathbf{n}=L\mathbf{z}$ where $\mathbf{z}\sim\mathcal{N}(0,I)$, and whitening is performed via $W$ such that $W\Sigma W^\top=I$. 

\subsection{Noise level parameter}

The parameter $\sigma_0$ controls the measurement-noise magnitude in DDRM's likelihood term. For simulated experiments, we use a consistent noise level $\sigma_0$ across all methods. For real-world experiments, we tune $\sigma_0$ separately for each baseline to ensure fair comparisons.

\subsection{Evaluation metrics}

We report three standard metrics, computed per image and then averaged: PSNR (dB) on $[0,1]$-normalized RGB images, SSIM on $[0,1]$ scale, and LPIPS using the VGG backbone. All metrics are computed on outputs mapped from the model's native $[-1,1]$ range back to $[0,1]$.

\subsection{Computational resources}
All experiments are executed on a single NVIDIA GeForce RTX~4090 GPU (24\,GB) running on an Exxact workstation and CUDA 13.0.

\section{DDRM Under Whitening}
\label{sec:supp-whitened-ddrm}
This section derives the DDRM updates when measurements are corrupted by correlated Gaussian noise and whitened prior to conditioning. Whitening modifies only the forward operator and measurements, and the diffusion schedule $\{\sigma_t\}_{t=0}^{T}$ and denoiser outputs remain unchanged. The resulting equations are DDRM's spectral one-dimensional Gaussians applied to the whitened problem (see\ Eqs.~(4)–(8) in~\cite{kawar2022ddrm}).

\subsection{Whitened measurement model}
We consider the linear observation model with correlated noise
\begin{equation}
\mathbf{y}=H\mathbf{x}_0+\mathbf{n},\qquad \mathbf{n}\sim\mathcal{N}(0,\sigma_y^2\Sigma).
\end{equation}
To restore independence, we left-precondition by the symmetric inverse square-root $W=\Sigma^{-1/2}$, obtaining
\begin{equation}
\tilde{\mathbf{y}}=W\mathbf{y},\qquad \tilde{H}=WH,\qquad \tilde{\mathbf{n}}=W\mathbf{n}\sim\mathcal{N}(0,\sigma_y^2 I).
\end{equation}
All conditioning steps are then carried out on $(\tilde{H},\tilde{\mathbf{y}})$ with i.i.d.\ noise. Let $\tilde{H}=\tilde{U}\tilde{S}\tilde{V}^\top$ denote the SVD. We work in spectral coordinates defined by
\begin{equation}
\bar{\tilde{\mathbf{x}}}_t=\tilde{V}^\top\mathbf{x}_t,\qquad
\bar{\tilde{\mathbf{y}}}=\tilde{S}^\dagger\tilde{U}^\top\tilde{\mathbf{y}}.
\end{equation}
The diffusion prior retains the usual Gaussian marginals
\begin{equation}
q(\mathbf{x}_t\mid\mathbf{x}_0)=\mathcal{N}(\mathbf{x}_0,\sigma_t^2 I),\qquad 0=\sigma_0<\cdots<\sigma_T,
\end{equation}
so whitening modifies only the measurement-consistency component of the update (replacing $s_i,\bar{y}^{(i)}$ by $\tilde{s}_i,\bar{\tilde{y}}^{(i)}$), while the prior side remains unchanged. For compact notation, we define
\begin{equation}
\delta_i\equiv \sigma_y/\tilde{s}_i,\qquad
\alpha_t\equiv \sqrt{1-\eta^2}\,\sigma_t,\qquad
\beta_t\equiv \alpha_t/\sigma_{t+1},
\end{equation}
and use shorthands $\xi_t^{(i)}:=\bar{\tilde{x}}_{t}^{(i)}$ and $\upsilon^{(i)}:=\bar{\tilde{y}}^{(i)}$ for spectral coordinates. Following DDRM, we assume $\sigma_T\ge \delta_i$ for all $\tilde{s}_i>0$ to ensure non-negative initial variance.

\subsection{Whitened DDRM updates}

\paragraph{Initialization ($t=T$).}
At the largest noise level, the posterior per spectral coordinate is Gaussian centered at the measurement component for observable coordinates ($\tilde{s}_i>0$), while unobservable coordinates ($\tilde{s}_i=0$) follow the unconditional prior:
\begin{equation}
q^{(T)}\!\big(\xi_T^{(i)}\mid \mathbf{x}_0,\tilde{\mathbf{y}}\big)=
\begin{cases}
\mathcal{N}\!\big(\upsilon^{(i)},\,\sigma_T^2-\delta_i^2\big), & \tilde{s}_i>0,\\
\mathcal{N}\!\big(\xi_0^{(i)},\,\sigma_T^2\big), & \tilde{s}_i=0,
\end{cases}
\end{equation}
\begin{equation}
p_\theta^{(T)}\!\big(\xi_T^{(i)}\mid \tilde{\mathbf{y}}\big)=
\begin{cases}
\mathcal{N}\!\big(\upsilon^{(i)},\,\sigma_T^2-\delta_i^2\big), & \tilde{s}_i>0,\\
\mathcal{N}\!\big(0,\,\sigma_T^2\big), & \tilde{s}_i=0.
\end{cases}
\end{equation}
These match DDRM's initializers with the substitutions $(s_i,\bar{y}^{(i)})\!\mapsto\!(\tilde{s}_i,\upsilon^{(i)})$.

\paragraph{Transitions ($t<T$): ground truth posterior $q$.}
Each step interpolates between the current clean estimate and the measurement across three scenarios. We abbreviate $q^{(t)}(\cdot| \mathbf{x}_{t+1},\mathbf{x}_0,\tilde{\mathbf{y}})$ as $q^{(t)}(\cdot|\cdot)$:
\begin{equation}
\begin{aligned}
q^{(t)}\!\big(\xi_t^{(i)}\big|\cdot\big)
&= \mathcal{N}\!\big(\,\xi_0^{(i)}+\beta_t(\xi_{t+1}^{(i)}-\xi_0^{(i)}),\, \eta^2\sigma_t^2\big), \\
&\quad \tilde{s}_i=0,
\end{aligned}
\end{equation}
\begin{equation}
\begin{aligned}
q^{(t)}\!\big(\xi_t^{(i)}\big|\cdot\big)
&= \mathcal{N}\!\big(\,\xi_0^{(i)}+\tfrac{\alpha_t}{\delta_i}(\,\upsilon^{(i)}-\xi_0^{(i)}\,),\, \eta^2\sigma_t^2\big), \\
&\quad 0<\sigma_t<\delta_i,
\end{aligned}
\end{equation}
\begin{equation}
\begin{aligned}
q^{(t)}\!\big(\xi_t^{(i)}\big|\cdot\big)
&= \mathcal{N}\!\big(\,(1-\eta_b)\xi_0^{(i)}+\eta_b\upsilon^{(i)},\, \sigma_t^2-\eta_b^2\delta_i^2\big), \\
&\quad \sigma_t\ge\delta_i.
\end{aligned}
\end{equation}
When a coordinate is unobserved ($\tilde{s}_i{=}0$), the step reduces to unconditional generation. When observed and diffusion noise dominates ($\sigma_t<\delta_i$), the update pulls toward the measurement with gain $\alpha_t/\delta_i$. Once measurement noise dominates ($\sigma_t\ge\delta_i$), the mean becomes a convex blend with weight $\eta_b$ and the variance reduces by $\eta_b^2\delta_i^2$ to preserve Gaussian marginals.

\paragraph{Transitions ($t<T$): sampling distribution $p_\theta$.}
Let $\mathbf{x}_{\theta,t}=f_\theta(\mathbf{x}_{t+1},t{+}1)$ and $\xi_{\theta,t}^{(i)}=(\tilde{V}^\top\mathbf{x}_{\theta,t})^{(i)}$. The sampling updates mirror the variational ones with $\xi_0^{(i)}$ replaced by $\xi_{\theta,t}^{(i)}$:
\begin{equation}
\begin{aligned}
p_\theta^{(t)}\!\big(\xi_t^{(i)}\big|\cdot\big)
&= \mathcal{N}\!\big(\,\xi_{\theta,t}^{(i)}+\beta_t(\xi_{t+1}^{(i)}-\xi_{\theta,t}^{(i)}),\, \eta^2\sigma_t^2\big), \\
&\quad \tilde{s}_i=0,
\end{aligned}
\end{equation}
\begin{equation}
\begin{aligned}
p_\theta^{(t)}\!\big(\xi_t^{(i)}\big|\cdot\big)
&= \mathcal{N}\!\big(\,\xi_{\theta,t}^{(i)}+\tfrac{\alpha_t}{\delta_i}(\,\upsilon^{(i)}-\xi_{\theta,t}^{(i)}\,),\, \eta^2\sigma_t^2\big), \\
&\quad 0<\sigma_t<\delta_i,
\end{aligned}
\end{equation}
\begin{equation}
\begin{aligned}
p_\theta^{(t)}\!\big(\xi_t^{(i)}\big|\cdot\big)
&= \mathcal{N}\!\big(\,(1-\eta_b)\xi_{\theta,t}^{(i)}+\eta_b\upsilon^{(i)},\, \sigma_t^2-\eta_b^2\delta_i^2\big), \\
&\quad \sigma_t\ge\delta_i.
\end{aligned}
\end{equation}
These are DDRM's per-coordinate sampling rules in whitened coordinates, with substitutions $(s_i,\bar{y}^{(i)})\!\mapsto\!(\tilde{s}_i,\upsilon^{(i)})$ and threshold $\sigma_y/s_i$ replaced by $\delta_i=\sigma_y/\tilde{s}_i$.

\paragraph{Interpretation of boundary conditions.}
After whitening, every coordinate has unit noise variance, but the forward operator is rescaled by $\Sigma^{-1/2}$. As a result, the transition between the “diffusion-dominated’’ and the “measurement-dominated’’ conditions occurs when the diffusion noise level $\sigma_t$ matches the effective measurement noise along each singular direction. This happens at $\sigma_t = \delta_i$, where $\delta_i$ is the standard deviation of the measurement noise in the whitened basis. Because whitening changes the singular values from $s_i$ to $\tilde{s}_i$, the boundary is determined by $\tilde{s}_i$, not the original $s_i$.

\section{CIN-D Capture Protocol}
\label{sec:capture}
\begin{figure*}[!tt]
  \centering
  \includegraphics[width=1.0\linewidth]{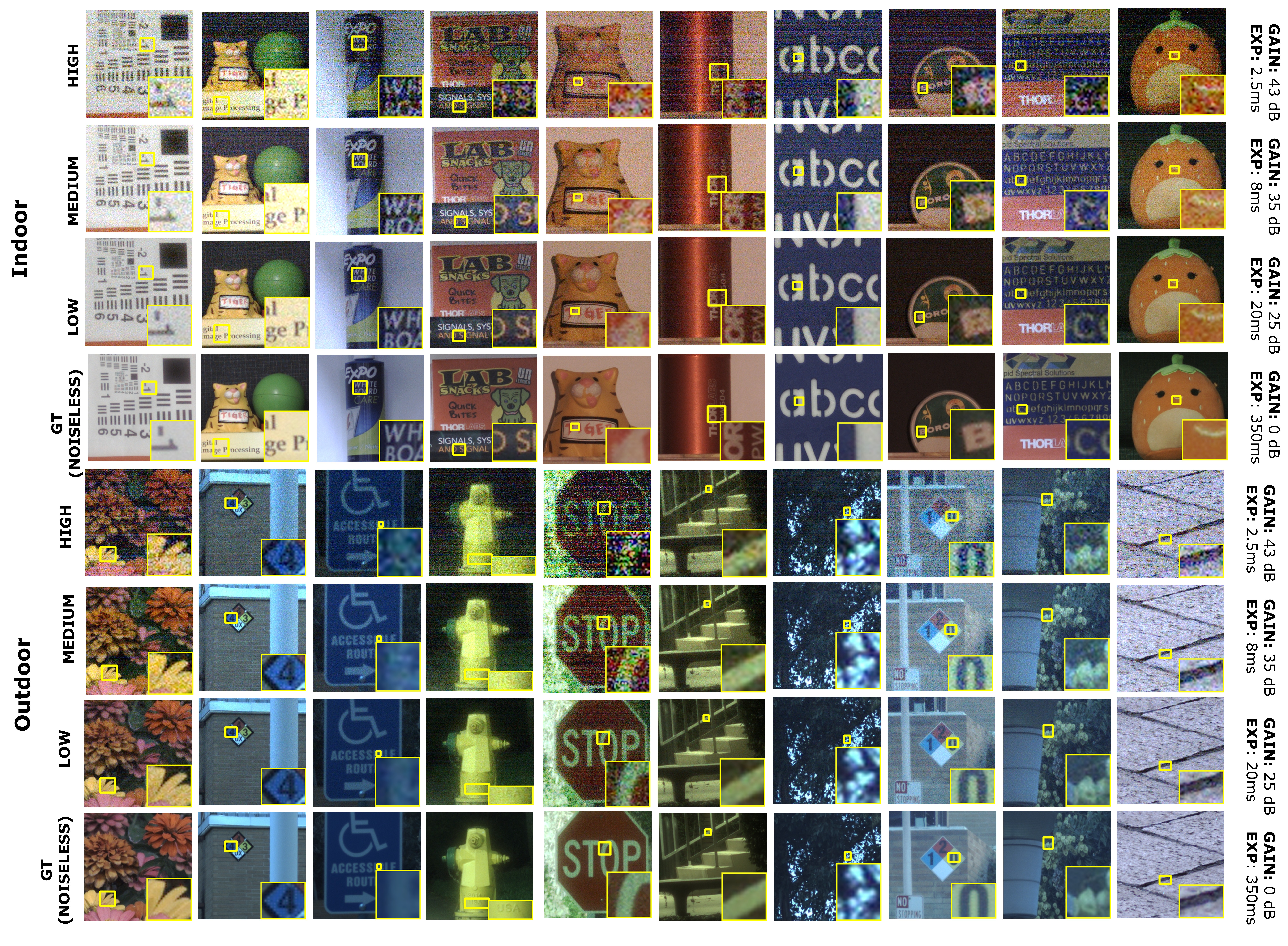}
    \caption{\textbf{CIN-D Dataset Examples.} Images of the same scenes captured at different noise levels using a FLIR Blackfly S BFS-U3-63S4C-C color camera. Noise levels decrease from top to bottom through controlled variations in sensor gain and exposure time, while scene content and viewing geometry remain fixed. This illustrates the progressive degradation in image quality that our dataset captures for benchmarking restoration algorithms under realistic correlated noise conditions.}
  \label{fig:dataset}
\end{figure*}

We collect the \emph{Correlated Image Noise Dataset} (CIN-D) using a rolling shutter machine-vision camera. The camera is configured to output 16-bit Bayer \texttt{BayerRG16} frames and, for convenience, 16-bit \texttt{Linear RGB} images. To reduce storage requirements and simplify subsequent noise modeling, we configure the camera's region of interest (ROI) to capture only a $256 \times 256$ center crop of the full-resolution sensor. This size matches the input resolution of the DDRM backbone, avoiding the need for resizing operations that would corrupt the natural noise structure. All gamma correction and automatic image-processing features (e.g., auto-exposure, auto-gain, auto-white-balance) are explicitly disabled to ensure a linear and stationary acquisition pipeline. We show representative examples in Fig.~\ref{fig:dataset} illustrating the diversity and the degradation seen in the same scenes due to higher correlated noise levels. 

\begin{table}[!tt]
\caption{Gain and exposure settings used to define the four noise levels in CIN-D. Higher gain and lower exposure yield stronger noise.}
\label{tab:cind_levels}
\centering
\resizebox{0.85\columnwidth}{!}{
\begin{tabular}{@{}lcc@{}}
\toprule
Noise level $\ell$ & Gain $g_\ell$ (dB) & Exposure $t_\ell$ (ms) \\
\midrule
zero   & 0.0  & 350  \\
low    & 25.0 & 20   \\
medium & 35.0 & 8    \\
high   & 43.0 & 2.5  \\
\bottomrule
\end{tabular}
}
\end{table}

\begin{table}[!tt]
    \centering
    \caption{Hardware components used for CIN-D data acquisition.}
    \label{tab:cind_parts}
    \begin{tabular}{p{0.22\linewidth} p{0.3\linewidth} p{0.3\linewidth}}
        \toprule
        Component & Model / Part no. & Key specifications \\
        \midrule
        Camera &
        FLIR Blackfly S BFS-U3-63S4C-C &
        6.3 MP color CMOS, Rolling Shutter\\
        Lens &
        Edmund Optics TECHSPEC HP, \#86-572 &
        25\,mm, $f/1.8$--$f/22$, C-mount\\
        Mounting (indoor) &
        Post Mounts &
        Fixed camera, repeatable alignment \\
        \bottomrule
    \end{tabular}
\end{table}

For each scene, we sequentially capture four images by switching between the noise levels summarized in Table~\ref{tab:cind_levels}. The gain and exposure combinations are carefully selected through trial and error to maintain approximately constant effective exposure, and hence overall scene brightness, across all four noise levels. This design keeps all acquisition variables fixed except for the noise-level control parameters, which is essential for studying denoising and restoration algorithms that assume identical underlying scene radiance. In the remainder of this section, we briefly describe the hardware setup (Table~\ref{tab:cind_parts}) and the acquisition strategies for indoor and outdoor capture.

\begin{figure}[!tt]
  \centering
  \includegraphics[width=0.9\linewidth]{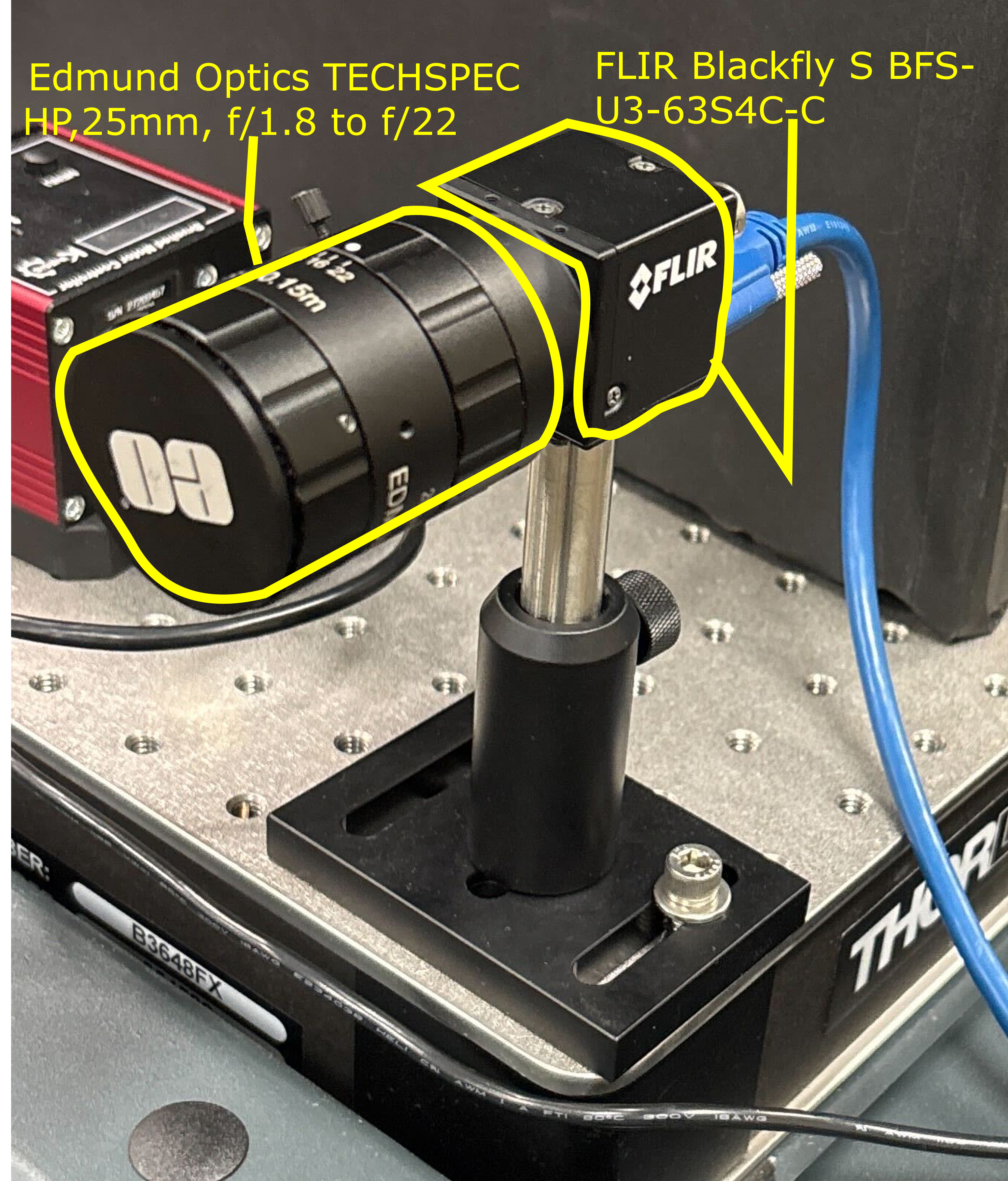}
  \caption{\textbf{Hardware setup.} Camera and lens assembly used for CIN-D acquisition. A FLIR Blackfly S BFS-U3-63S4C-C camera is paired with a fixed-focal-length C-mount lens and rigidly mounted on an optical bench for indoor scene capture.}

  \label{fig:hardware}
\end{figure}

\paragraph{Hardware setup and acquisition strategy}

\begin{figure}[!tt]
  \centering
  \includegraphics[width=1.0\linewidth]{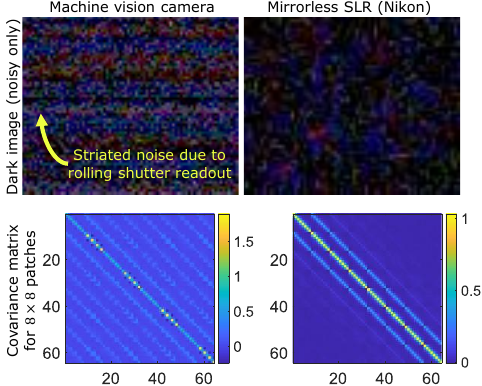}
    \caption{\textbf{Comparison of dark-frame noise in a machine-vision rolling-shutter camera and a mirrorless SLR.} Top row: Dark frames showing pronounced horizontal striations in the machine-vision camera due to rolling-shutter readout, versus more homogeneous noise in the Nikon mirrorless camera. Bottom row: Noise covariance matrices of $8\times8$ luminance patches, showing anisotropic correlations for the rolling-shutter sensor, compared to a weakly correlated noise profile for the mirrorless SLR.}
  \label{fig:noise-covariance-matrix}
  \end{figure}

We acquire all images in CIN-D using a FLIR Blackfly S BFS-U3-63S4C-C color rolling-shutter camera coupled with an Edmund Optics TECHSPEC HP 25\,mm C-mount lens as shown in Fig.~\ref{fig:hardware}. This lens delivers well-corrected, low-distortion imaging with an adjustable aperture that allows us to control optical throughput and prevent saturation under different lighting conditions. For indoor captures, we mount the camera rigidly on an optical bench to maintain consistent viewing geometry while varying only the target and illumination across scenes. Indoor targets remain static under approximately uniform room lighting or controlled illumination, and we adjust the lens $f$-number as necessary to keep sensor response within a non-saturated, signal-dominated regime at fixed gain and exposure settings. For outdoor captures, we follow the same principle of imaging static content while ensuring the camera remains completely stationary throughout acquisition by securing it firmly to a rigid surface using a tripod laptop stand and duct tape. Under both indoor and outdoor conditions, we carefully avoid moving objects, rapid illumination changes, and strong specular highlights so that dataset variation is dominated by our controlled changes in exposure, gain, and noise statistics rather than by scene dynamics. We sequentially vary the noise level using a custom acquisition script that leverages the PySpin (Spinnaker) SDK to access, acquire, and process camera data programmatically. This approach is essential for benchmarking denoising and restoration methods under identical scene radiance and viewing geometry but with progressively challenging noise conditions. The 0\,dB gain frame serves as a substitute ground-truth clean image, enabling quantitative evaluation of denoising performance on the corresponding noisy images.

\paragraph{Dark frame acquisition}
We capture dark frames to characterize sensor noise and estimate the per-noise-level spatial covariance matrix. The process mirrors the actual acquisition strategy but with the lens cap in place to block all light from reaching the sensor. Dark frames use exactly the same $(t_\ell, g_\ell)$ settings as specified in Table~\ref{tab:cind_levels} and are stored in parallel directory trees using the same logging format. As shown in Fig.~\ref{fig:noise-covariance-matrix}, computing the noise covariance matrix allows us to quantify the strength, spatial extent, and anisotropy of correlated noise at each level, which informs the construction of our whitening transform.

\begin{figure*}[!tt]
  \centering
  \includegraphics[width=0.95\linewidth]{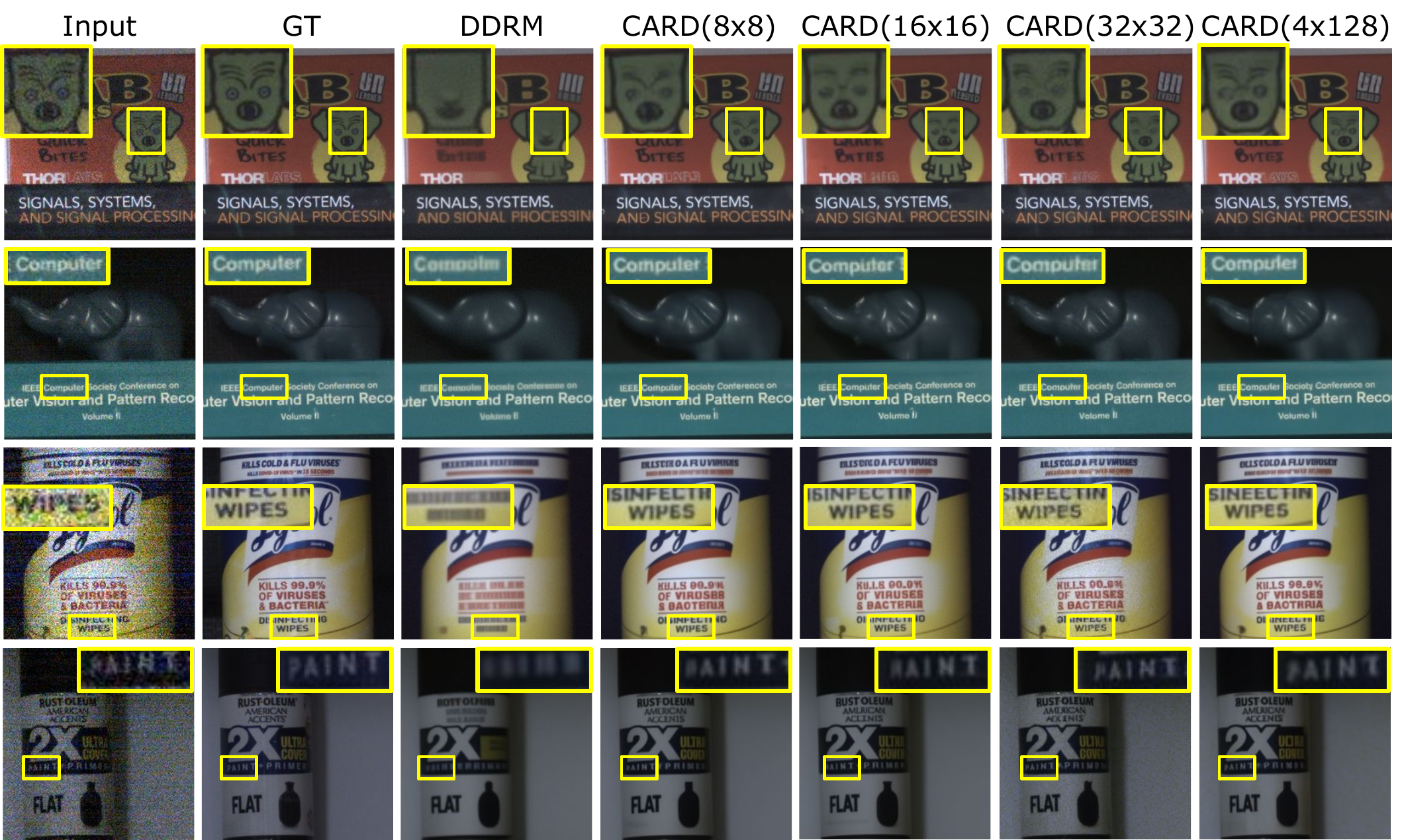}
    \caption{\textbf{Effect of patch size on CARD denoising quality.} Top two rows show low-noise images where smaller patches ($8{\times}8$, $16{\times}16$) produce slightly clearer outputs. Bottom two rows show high-noise images where larger and anisotropic patches ($32{\times}32$, $4{\times}128$) yield improved visual quality by better capturing spatial noise correlations.}
  \label{fig:supp-patch-size-figure}
\end{figure*}

\begin{table}[!tt]
\caption{Effect of patch size on denoising CIN-D indoor images using CARD. Values are PSNR/SSIM/LPIPS (P/S/L). Each patch size corresponds to the local neighborhood used for covariance estimation and whitening.}
\label{tab:patchsize_real}
\centering
\resizebox{0.9\columnwidth}{!}{
\begin{tabular}{@{}lcc@{}}
\toprule
Patch Size & Low Noise (P/S/L) & High Noise (P/S/L) \\
\midrule
$8 \times 8$   & 39.8 / 0.96 / 0.17 & 33.1 / 0.87 / 0.29 \\
$16 \times 16$ & 39.8 / 0.96 / 0.17 & 33.2 / 0.87 / 0.27 \\
$32 \times 32$ & 39.2 / 0.94 / 0.15 & 33.3 / 0.88 / 0.27 \\
$4 \times 128$ & 39.8 / 0.96 / 0.16 &33.4 / 0.88 / 0.25 \\
\bottomrule
\end{tabular}
}
\end{table}

\section{Patch Size for Covariance Estimation}
\label{sec:ablation-patchsize}
The patch size used for covariance estimation has a modest yet visible effect on denoising quality, as shown in Table~\ref{tab:patchsize_real} and Figure~\ref{fig:supp-patch-size-figure}. At low noise levels, all patch sizes achieve comparable PSNR ($\sim$39.8\,dB) and SSIM ($\sim$0.96), indicating that CARD remains stable across spatial scales. However, perceptual quality varies slightly, with the $32{\times}32$ patch achieving the lowest LPIPS (0.15). Qualitatively, as seen in the top two rows of Figure~\ref{fig:supp-patch-size-figure}, smaller patch sizes such as $8{\times}8$ and $16{\times}16$ produce slightly clearer outputs at low noise, likely because local noise statistics are well-captured within compact neighborhoods when noise is mild.

At high noise levels, the elongated $4{\times}128$ patch attains the highest PSNR (33.4\,dB) and lowest LPIPS (0.25) while maintaining high SSIM (0.88), suggesting it better captures the horizontal correlations characteristic of rolling-shutter readout. As illustrated in the bottom two rows of Figure~\ref{fig:supp-patch-size-figure}, larger patches yield improved visual quality under severe noise, presumably because they aggregate sufficient samples to reliably estimate the stronger spatial correlations present at high gain settings. CARD proves robust to patch size choice, with $8{\times}8$ patches offering a good balance between computational efficiency and performance.

\section{Additional Experiments}

\subsection{Experiments with simulated correlated noise}
We evaluate CARD on denoising, deblurring, and super-resolution tasks across ImageNet~\cite{deng2009imagenet}, LSUN-Bed, and LSUN-Cat~\cite{yu2015lsun} datasets under simulated correlated noise.

\paragraph{Denoising results.}
We evaluate CARD on image denoising under simulated correlated noise across three datasets: ImageNet, LSUN-Bed, and LSUN-Cat. The quantitative results are summarized in Tables~\ref{tab:deno-imagenet}, \ref{tab:deno-lsunbed}, and \ref{tab:deno-lsuncat} for noise levels $\sigma_0 \in \{0.1, 0.5, 0.9\}$. Across all datasets and noise levels, CARD achieves the highest PSNR and SSIM, while also obtaining the lowest LPIPS. Learning-based denoisers trained under the i.i.d.\ noise assumption (Restormer, DnCNN, Noise2Info, PCST) experience substantial performance degradation as noise becomes more correlated or more severe. Classical and diffusion-based i.i.d.\ methods (BM3D, DDNM, DDRM) remain competitive at low noise levels but deteriorate rapidly for large $\sigma_0$. In contrast, CARD continues to outperform all baselines by a significant margin. Learning-based models designed specifically for correlated noise (APRRD-BSN and APRRD-NBSN) offer more robustness than i.i.d.\ baselines, yet they consistently fall short of CARD on all metrics.

\begin{figure*}[!tt]
  \centering
  \includegraphics[width=0.99\linewidth]{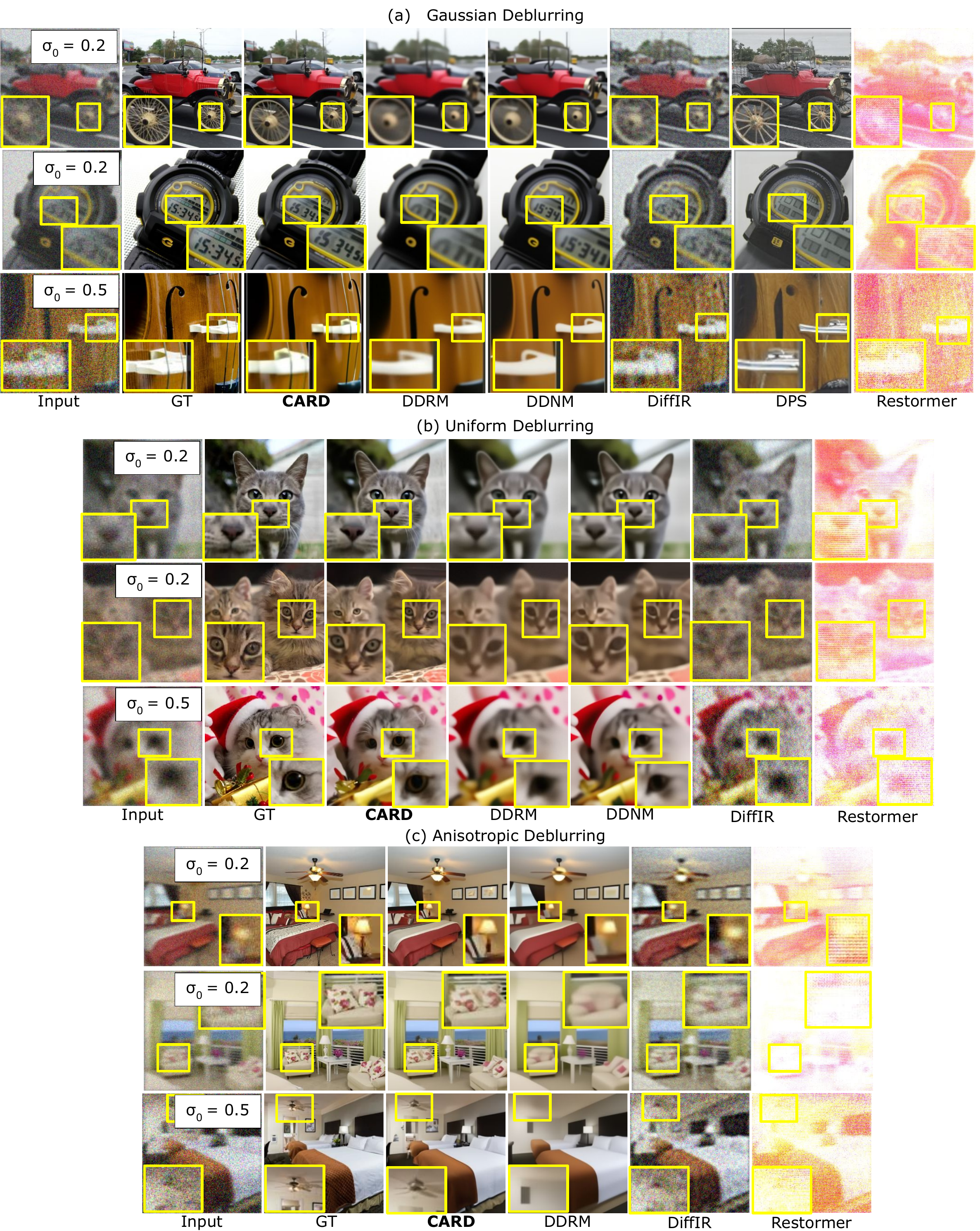}
    \caption{\textbf{Qualitative comparison of Gaussian, Uniform, and Anisotropic deblurring} at varying noise levels ($\sigma_0=0.2, 0.5$) on ImageNet (top three rows), LSUN-Cat (middle three rows), and LSUN-Bedroom (bottom three rows). The boxed regions highlight areas of interest where differences are most apparent. CARD outperforms other baselines while preserving fine details.}
  \label{fig:supp-all-figure}
\end{figure*}

\paragraph{Deblurring results.}
We evaluate the effectiveness of CARD under simulated correlated noise across three deblurring settings: Gaussian, Uniform, and Anisotropic blur. Figure~\ref{fig:supp-all-figure} presents qualitative comparisons on ImageNet, LSUN-Cat, and LSUN-Bed at two noise levels ($\sigma_0{=}0.2$ and $0.5$). Across all datasets and blur types, CARD produces sharper reconstructions and preserves fine structures more effectively than existing diffusion-based and learning-based approaches. Tables~\ref{tab:deblur-imagenet}, \ref{tab:deblur-lsunbed}, and~\ref{tab:deblur-lsuncat} report quantitative results. CARD achieves the highest PSNR and SSIM, while also obtaining the lowest LPIPS across all combinations of blur type, dataset, and noise level. 

\paragraph{Super-resolution results.}
We evaluate CARD for $2\times$ and $4\times$ super-resolution under medium and high correlated noise. Tables~\ref{tab:sr-imagenet-sup}, \ref{tab:sr-lsunbed}, and \ref{tab:sr-lsuncat} summarize results on ImageNet, LSUN-Bed, and LSUN-Cat. Across all datasets, scales, and noise levels, CARD consistently achieves the highest PSNR and SSIM, while obtaining the lowest LPIPS. The gains are particularly significant at higher noise levels and for the more challenging $4\times$ setting, where other methods struggle to maintain stability.

\begin{table*}[th]
\caption{\textbf{Denoising results on ImageNet} across low (0.1), medium (0.5), and high (0.9) noise levels. We report PSNR/SSIM/LPIPS as P/S/L. Best and second‐best results are marked in bold and underlined, respectively. CARD achieves the best performance across all noise levels.}
\label{tab:deno-imagenet}
\centering
\resizebox{0.8\linewidth}{!}{
\begin{tabular}{@{}llccc@{}}
\toprule
\multirow{2}{*}{Category} &
\multirow{2}{*}{Model} &
\multicolumn{3}{c}{Noise Level $\sigma_0$} \\
 & & 0.1(P/S/L) & 0.5(P/S/L) & 0.9(P/S/L) \\
\midrule
\multirow{4}{*}{Learning-based (i.i.d.)}
& Restormer~\cite{zamir2022restormer}   & 10.5/0.40/0.78 & 10.3/0.26/0.96 & 10.1/0.16/1.14 \\
& DnCNN~\cite{zhang2017dncnn}      & 22.6/0.87/0.24 & 20.4/0.61/0.46 & 18.1/0.40/0.66 \\
& Noise2Info~\cite{wang2023noise2info} & 25.2/0.80/0.24 & 24.6/0.74/\underline{0.26} & \underline{23.3}/0.65/\underline{0.29} \\
& PCST~\cite{vaksman2023pcst}        & 25.6/0.71/0.36 & 19.8/0.48/0.63 & 16.6/0.36/0.79 \\
\midrule
\multirow{3}{*}{Non-learning (i.i.d.)}
& BM3D~\cite{dabov2007bm3d}        & 30.1/0.78/\underline{0.11} & \underline{25.8}/0.71/0.33 & 16.6/0.36/0.79 \\
& DDNM~\cite{wang2022ddnm}        & 28.1/0.83/0.25 & 16.3/0.24/0.57 & 12.3/0.10/0.68 \\
& DDRM~\cite{kawar2022ddrm}        & \underline{31.0/0.92}/0.14 & 24.8/\underline{0.79}/0.33 & 22.7/\underline{0.74}/0.40 \\
\midrule
\multirow{2}{*}{Learning-based (correlated)}
& APRRD-BSN~\cite{kim2025apr}   & 23.9/0.72/0.25 & 22.0/0.60/0.38 & 20.1/0.52/0.51 \\
& APRRD-NBSN~\cite{kim2025apr}  & 24.9/0.76/0.31 & 22.4/0.63/0.45 & 20.5/0.54/0.54 \\
\midrule
\textbf{Ours}
& \textbf{CARD} & \textbf{34.0}/\textbf{0.96}/\textbf{0.07} & \textbf{29.1}/\textbf{0.90}/\textbf{0.15} & \textbf{26.7}/\textbf{0.86}/\textbf{0.22} \\
\bottomrule
\end{tabular}}
\end{table*}

\begin{table*}[th]
\caption{\textbf{Denoising results on LSUN-Bed} across low (0.1), medium (0.5), and high (0.9) noise levels. We report PSNR/SSIM/LPIPS as P/S/L. Best and second‐best results are marked in bold and underlined, respectively. CARD achieves the best performance across all noise levels.}
\label{tab:deno-lsunbed}
\centering
\resizebox{0.8\linewidth}{!}{
\begin{tabular}{@{}llccc@{}}
\toprule
\multirow{2}{*}{Category} &
\multirow{2}{*}{Model} &
\multicolumn{3}{c}{Noise Level $\sigma_0$} \\
 & & 0.1(P/S/L) & 0.5(P/S/L) & 0.9(P/S/L) \\
\midrule
\multirow{4}{*}{Learning-based (i.i.d.)}
& Restormer~\cite{zamir2022restormer}   & 10.0/0.41/0.76 & 10.1/0.29/0.95 & 10.2/0.18/1.14 \\
& DnCNN~\cite{zhang2017dncnn}       & 23.2/0.90/0.20 & 21.1/0.63/0.45 & 18.7/0.41/0.67 \\
& Noise2Info~\cite{wang2023noise2info}  & 26.3/0.83/0.21 & 25.7/0.78/\underline{0.23} & \underline{24.1}/0.69/\underline{0.29} \\
& PCST~\cite{vaksman2023pcst}        & 25.8/0.76/0.33 & 20.2/0.53/0.65 & 17.0/0.40/0.81 \\
\midrule
\multirow{3}{*}{Non-learning (i.i.d.)}
& BM3D~\cite{dabov2007bm3d}        & 30.1/0.74/0.14 & \underline{26.1}/0.73/0.31 & 22.5/0.61/0.50 \\
& DDNM~\cite{wang2022ddnm}        & 28.5/0.83/0.23 & 16.3/0.24/0.55 & 12.2/0.10/0.67 \\
& DDRM~\cite{kawar2022ddrm}        & \underline{32.2/0.95/0.10} & 26.0/\underline{0.85}/0.25 & 23.9/\underline{0.81}/0.30 \\
\midrule
\multirow{2}{*}{Learning-based (correlated)}
& APRRD-BSN~\cite{kim2025apr}   & 24.5/0.75/0.27 & 22.5/0.65/0.40 & 20.8/0.57/0.53 \\
& APRRD-NBSN~\cite{kim2025apr}  & 25.8/0.79/0.31 & 23.1/0.67/0.47 & 21.2/0.59/0.58 \\
\midrule
\multirow{1}{*}{Ours}
& \textbf{CARD} & \textbf{35.1}/\textbf{0.97}/\textbf{0.05} & \textbf{29.6}/\textbf{0.92}/\textbf{0.12} & \textbf{27.2}/\textbf{0.89}/\textbf{0.17} \\
\bottomrule
\end{tabular}}
\end{table*}

\begin{table*}[th]
\caption{\textbf{Denoising results on LSUN-Cat} across low (0.1), medium (0.5), and high (0.9) noise levels. We report PSNR/SSIM/LPIPS as P/S/L. Best and second‐best results are marked in bold and underlined, respectively. CARD achieves the best performance across all noise levels.}
\label{tab:deno-lsuncat}
\centering
\resizebox{0.8\linewidth}{!}{
\begin{tabular}{@{}llccc@{}}
\toprule
\multirow{2}{*}{Category} &
\multirow{2}{*}{Model} &
\multicolumn{3}{c}{Noise Level $\sigma_0$} \\
 & & 0.1(P/S/L) & 0.5(P/S/L) & 0.9(P/S/L) \\
\midrule
\multirow{4}{*}{Learning-based (i.i.d.)}
& Restormer~\cite{zamir2022restormer}   & 10.6/0.38/0.80 & 10.4/0.25/0.98 & 10.1/0.16/1.16 \\
& DnCNN~\cite{zhang2017dncnn}       & 24.0/0.90/0.20 & 21.3/0.61/0.45 & 18.5/0.39/0.67 \\
& Noise2Info~\cite{wang2023noise2info}  & 25.5/0.81/0.25 & 25.0/0.76/\underline{0.26} & 23.7/0.68/\underline{0.29} \\
& PCST~\cite{vaksman2023pcst}       & 26.7/0.75/0.35 & 20.2/0.52/0.61 & 16.8/0.39/0.77 \\
\midrule
\multirow{3}{*}{Non-learning (i.i.d.)}
& BM3D~\cite{dabov2007bm3d}        & 30.1/0.76/\underline{0.14} & \underline{26.1}/0.72/0.34 & 22.4/0.61/0.51 \\
& DDNM~\cite{wang2022ddnm}        & 28.3/0.82/0.27 & 16.5/0.24/0.57 & 12.4/0.10/0.68 \\
& DDRM~\cite{kawar2022ddrm}        & \underline{31.6/0.93}/0.15 & \underline{26.1/0.83}/0.30 & \underline{24.1/0.80}/0.35 \\
\midrule
\multirow{2}{*}{Learning-based (correlated)}
& APRRD-BSN~\cite{kim2025apr}   & 24.2/0.76/0.25 & 22.7/0.65/0.35 & 20.9/0.57/0.47 \\
& APRRD-NBSN~\cite{kim2025apr}  & 25.5/0.80/0.31 & 23.1/0.67/0.44 & 21.2/0.59/0.52 \\
\midrule
\multirow{1}{*}{Ours}
& \textbf{CARD} & \textbf{34.3}/\textbf{0.96}/\textbf{0.07} & \textbf{29.6}/\textbf{0.91}/\textbf{0.16} & \textbf{27.3}/\textbf{0.88}/\textbf{0.22} \\
\bottomrule
\end{tabular}}
\end{table*}

\begin{table*}[th]
\caption{\textbf{Deblurring results on ImageNet} across medium (0.2), and high (0.5) noise levels. We report PSNR/SSIM/LPIPS as P/S/L. Best and second‐best results are marked in bold and underlined, respectively. CARD achieves the best performance across all noise levels.}
\label{tab:deblur-imagenet}
\centering
\resizebox{0.95\linewidth}{!}{
\begin{tabular}{@{}lcccccc@{}}
\toprule
\multirow{2}{*}{Model} &
\multicolumn{2}{c}{Gaussian Deblur} &
\multicolumn{2}{c}{Anisotropic Deblur} &
\multicolumn{2}{c}{Uniform Deblur} \\
& 0.2 (P/S/L) & 0.5 (P/S/L) & 0.2 (P/S/L) & 0.5 (P/S/L) & 0.2 (P/S/L) & 0.5 (P/S/L) \\
\midrule
Restormer~\cite{zamir2022restormer} & 7.8/0.26/0.94 & 8.0/0.16/1.08 & 8.5/0.25/0.93 & 8.5/0.16/1.05 & 8.1/0.23/0.97 & 8.2/0.15/1.09 \\
DPS~\cite{chung2023dps}  & 19.4/0.53/0.41 & 19.3/0.51/0.46 & -- & -- & -- & -- \\
DiffIR~\cite{xia2023diffir}     & 20.1/0.43/0.65 & 17.8/0.19/0.98 & 11.0/0.19/0.92 & 10.7/0.08/1.10 & 20.7/0.37/0.80 & 18.2/0.15/1.05 \\
DDNM~\cite{wang2022ddnm}    & 21.8/0.53/0.45 & 15.9/0.21/0.60 & -- & -- & 21.2/0.50/0.49 & 15.8/0.20/0.62 \\
DDRM~\cite{kawar2022ddrm}       & \underline{24.3/0.77/0.36} & \underline{22.2/0.72/0.42} & \underline{24.1/0.76/0.37} & \underline{22.4/0.72/0.42} & \underline{22.9/0.72/0.41} & \underline{21.4/0.69/0.46} \\
\textbf{CARD (ours)} & \textbf{26.6}/\textbf{0.84}/\textbf{0.23} & \textbf{24.4}/\textbf{0.79}/\textbf{0.30} & \textbf{25.9}/\textbf{0.82}/\textbf{0.24} & \textbf{23.9}/\textbf{0.77}/\textbf{0.30} & \textbf{24.1}/\textbf{0.77}/\textbf{0.29} & \textbf{22.3}/\textbf{0.72}/\textbf{0.33} \\
\bottomrule
\end{tabular}}
\end{table*}

\begin{table*}[th]
\caption{\textbf{Deblurring results on LSUN-Bed} across medium (0.2), and high (0.5) noise levels. We report PSNR/SSIM/LPIPS as P/S/L. Best and second‐best results are marked in bold and underlined, respectively. CARD achieves the best performance across all noise levels.}
\label{tab:deblur-lsunbed}
\centering
\resizebox{0.95\linewidth}{!}{
\begin{tabular}{@{}lcccccc@{}}
\toprule
\multirow{2}{*}{Model} & \multicolumn{2}{c}{Gaussian Deblur} & \multicolumn{2}{c}{Anisotropic Deblur} & \multicolumn{2}{c}{Uniform Deblur} \\
& 0.2 (P/S/L) & 0.5 (P/S/L) & 0.2 (P/S/L) & 0.5 (P/S/L) & 0.2 (P/S/L) & 0.5 (P/S/L) \\
\midrule
Restormer~\cite{zamir2022restormer}   & 8.4/0.33/0.91  & 8.6/0.20/1.07  & 8.8/0.31/0.89  & 8.9/0.22/1.02  & 8.5/0.31/0.93  & 8.7/0.20/1.06 \\
DPS~\cite{chung2023dps}      & 19.7/0.58/0.40 & 19.3/0.56/\underline{0.46} & -- & -- & -- & -- \\
DiffIR~\cite{xia2023diffir}      & 20.5/0.43/0.72 & 18.1/0.18/1.05 & 22.6/0.44/0.71 & 19.3/0.19/1.01 & 20.9/0.38/0.85 & 18.4/0.15/1.11 \\
DDNM~\cite{wang2022ddnm}        & 22.0/0.54/0.42 & 15.9/0.21/0.58 & -- & -- & 21.5/0.51/0.45 & 15.9/0.20/0.59 \\
DDRM~\cite{kawar2022ddrm}       & \underline{25.1/0.82/0.28} & \underline{23.0/0.78}/\textbf{0.34} & \underline{24.8/0.81/0.29} & \underline{23.1/0.78/0.33} &\underline{23.7/0.78/0.32} & \underline{22.2/0.75/0.36} \\
\textbf{CARD (ours)} & \textbf{26.6}/\textbf{0.86}/\textbf{0.19} & \textbf{24.5}/\textbf{0.81}/\textbf{0.34} & \textbf{25.9}/\textbf{0.82}/\textbf{0.24} & \textbf{24.0}/\textbf{0.80}/\textbf{0.25} & \textbf{24.4}/\textbf{0.80}/\textbf{0.24} & \textbf{22.6}/\textbf{0.76}/\textbf{0.29} \\
\bottomrule
\end{tabular}}
\end{table*}

\begin{table*}[th]
\caption{\textbf{Deblurring results on LSUN-Cat} across medium (0.2), and high (0.5) noise levels. We report PSNR/SSIM/LPIPS as P/S/L. Best and second‐best results are marked in bold and underlined, respectively. CARD achieves the best performance across all noise levels.}
\label{tab:deblur-lsuncat}
\centering
\resizebox{0.95\linewidth}{!}{
\begin{tabular}{@{}lcccccc@{}}
\toprule
\multirow{2}{*}{Model} & \multicolumn{2}{c}{Gaussian Deblur} & \multicolumn{2}{c}{Anisotropic Deblur} & \multicolumn{2}{c}{Uniform Deblur} \\
& 0.2 (P/S/L) & 0.5 (P/S/L) & 0.2 (P/S/L) & 0.5 (P/S/L) & 0.2 (P/S/L) & 0.5 (P/S/L) \\
\midrule
Restormer~\cite{zamir2022restormer}              & 7.9/0.26/0.94  & 8.0/0.16/1.08  & 8.5/0.25/0.92  & 8.5/0.17/1.04  & 8.5/0.25/0.92  & 8.5/0.17/1.04 \\
DPS~\cite{chung2023dps}                    & 19.1/0.57/0.40              & 18.9/0.54/0.44              & --              & --              & --              & -- \\
DiffIR~\cite{xia2023diffir}                 & 19.8/0.44/0.68 & 17.7/0.19/1.02 & 22.5/0.46/0.67 & 19.2/0.21/0.98 & 20.6/0.39/0.82 & 18.2/0.16/1.08 \\
DDNM~\cite{wang2022ddnm}                   & 22.3/0.55/0.45 & 16.2/0.22/0.60              & --              & --              & 21.8/0.52/0.48              & 16.0/0.21/0.61 \\
DDRM~\cite{kawar2022ddrm}                   & 25.5/0.82/0.32 & \underline{23.2/0.77/0.38} & \underline{25.2/0.81/0.33} & \underline{23.3/0.77/0.38} & \underline{24.0/0.78/0.37} & \underline{22.4/0.75/0.41} \\
\textbf{CARD (ours)}  & \textbf{27.6}/\textbf{0.87}/\textbf{0.22} & \textbf{25.2}/\textbf{0.82}/\textbf{0.28} & \textbf{26.8}/\textbf{0.85}/\textbf{0.23} & \textbf{24.7}/\textbf{0.81}/\textbf{0.28} & \textbf{25.1}/\textbf{0.81}/\textbf{0.26} & \textbf{23.2}/\textbf{0.77}/\textbf{0.31} \\
\bottomrule
\end{tabular}}
\end{table*}

\begin{table*}[!tt]
\caption{\textbf{Super-resolution results on ImageNet} across $2\times$ and $4\times$ upscaling, medium (0.2) and high (0.5) noise levels. We report PSNR/SSIM/LPIPS as P/S/L. Best and second-best results are marked in bold and underlined, respectively. CARD achieves the best performance for both 2$\times$ and 4$\times$ upscaling, and at both noise levels.}
\label{tab:sr-imagenet-sup}
\centering
\resizebox{0.7\linewidth}{!}{
\begin{tabular}{@{}lcccc@{}}
\toprule
\multirow{2}{*}{Model} & \multicolumn{2}{c}{SR2} & \multicolumn{2}{c}{SR4} \\
& 0.2 (P/S/L) & 0.5 (P/S/L) & 0.2 (P/S/L) & 0.5 (P/S/L) \\
\midrule
DiffIR~\cite{xia2023diffir}        & \underline{25.7}/0.71/0.34 &\underline{23.9}/0.65/0.42 & \underline{23.8}/0.63/0.47 & \underline{23.0}/0.53/0.60 \\
DDRM~\cite{kawar2022ddrm}          & \underline{25.7/0.81/0.32} & 22.7/\underline{0.74/0.41} & 22.9/\underline{0.74/0.40} & 20.4/\underline{0.69/0.46} \\
\textbf{CARD (ours)} & \textbf{28.0}/\textbf{0.86}/\textbf{0.20} & \textbf{25.1}/\textbf{0.77}/\textbf{0.33} & \textbf{25.0}/\textbf{0.78}/\textbf{0.30} & \textbf{23.1}/\textbf{0.71}/\textbf{0.40} \\
\bottomrule
\end{tabular}}
\end{table*}

\begin{table*}[!tt]
\caption{\textbf{Super-resolution results on LSUN-Bed} across $2\times$ and $4\times$ upscaling, medium (0.2) and high (0.5) noise levels. We report PSNR/SSIM/LPIPS as P/S/L. Best and second-best results are marked in bold and underlined, respectively. CARD achieves the best performance for both 2$\times$ and 4$\times$ upscaling, and at both noise levels.}
\label{tab:sr-lsunbed}
\centering
\resizebox{0.7\linewidth}{!}{
\begin{tabular}{@{}lcccc@{}}
\toprule
\multirow{2}{*}{Model} & \multicolumn{2}{c}{SR2} & \multicolumn{2}{c}{SR4} \\
& 0.2 (P/S/L) & 0.5 (P/S/L) & 0.2 (P/S/L) & 0.5 (P/S/L) \\
\midrule
DiffIR~\cite{xia2023diffir}        & \underline{26.8}/0.72/0.32 & \underline{23.6}/0.47/0.62 & \underline{24.5}/0.67/0.40 & \underline{23.7}/0.56/0.56 \\
DDRM~\cite{kawar2022ddrm}          & 26.1/\underline{0.85/0.24} & 23.5/\underline{0.80/0.32} & 23.1/\underline{0.78/0.31} & 20.9/\underline{0.74/0.37} \\
\textbf{CARD (ours)} & \textbf{28.9}/\textbf{0.90}/\textbf{0.14} & \textbf{27.0}/\textbf{0.86}/\textbf{0.22} & \textbf{25.9}/\textbf{0.83}/\textbf{0.24} & \textbf{25.0}/\textbf{0.81}/\textbf{0.29} \\
\bottomrule
\end{tabular}}
\end{table*}

\begin{table*}[!tt]
\caption{\textbf{Super-resolution results on LSUN-Cat} across $2\times$ and $4\times$ upscaling, medium (0.2) and high (0.5) noise levels. We report PSNR/SSIM/LPIPS as P/S/L. Best and second-best results are marked in bold and underlined, respectively. CARD achieves the best performance for both 2$\times$ and 4$\times$ upscaling, and at both noise levels.}
\label{tab:sr-lsuncat}
\centering
\resizebox{0.7\linewidth}{!}{
\begin{tabular}{@{}lcccc@{}}
\toprule
\multirow{2}{*}{Model} & \multicolumn{2}{c}{SR2} & \multicolumn{2}{c}{SR4} \\
& 0.2 (P/S/L) & 0.5 (P/S/L) & 0.2 (P/S/L) & 0.5 (P/S/L) \\
\midrule
DiffIR~\cite{xia2023diffir}        &\underline{27.6}/0.75/0.34 & \underline{24.0}/0.50/0.64 & \underline{25.4}/0.69/0.48 & \underline{24.3}/0.58/0.63 \\
DDRM~\cite{kawar2022ddrm}          & 26.6/\underline{0.85/0.30} & 23.7/\underline{0.79/0.37} & 23.7/\underline{0.79/0.36} & 21.1/\underline{0.75/0.41} \\
\textbf{CARD (ours)} & \textbf{29.1}/\textbf{0.88}/\textbf{0.19} & \textbf{26.9}/\textbf{0.83}/\textbf{0.27} & \textbf{26.5}/\textbf{0.83}/\textbf{0.27} & \textbf{25.2}/\textbf{0.80}/\textbf{0.33} \\
\bottomrule
\end{tabular}}
\end{table*}

\subsection{Experiments on CIN-D}
\begin{figure*}[!tt]
  \centering
  \includegraphics[width=1.0\linewidth]{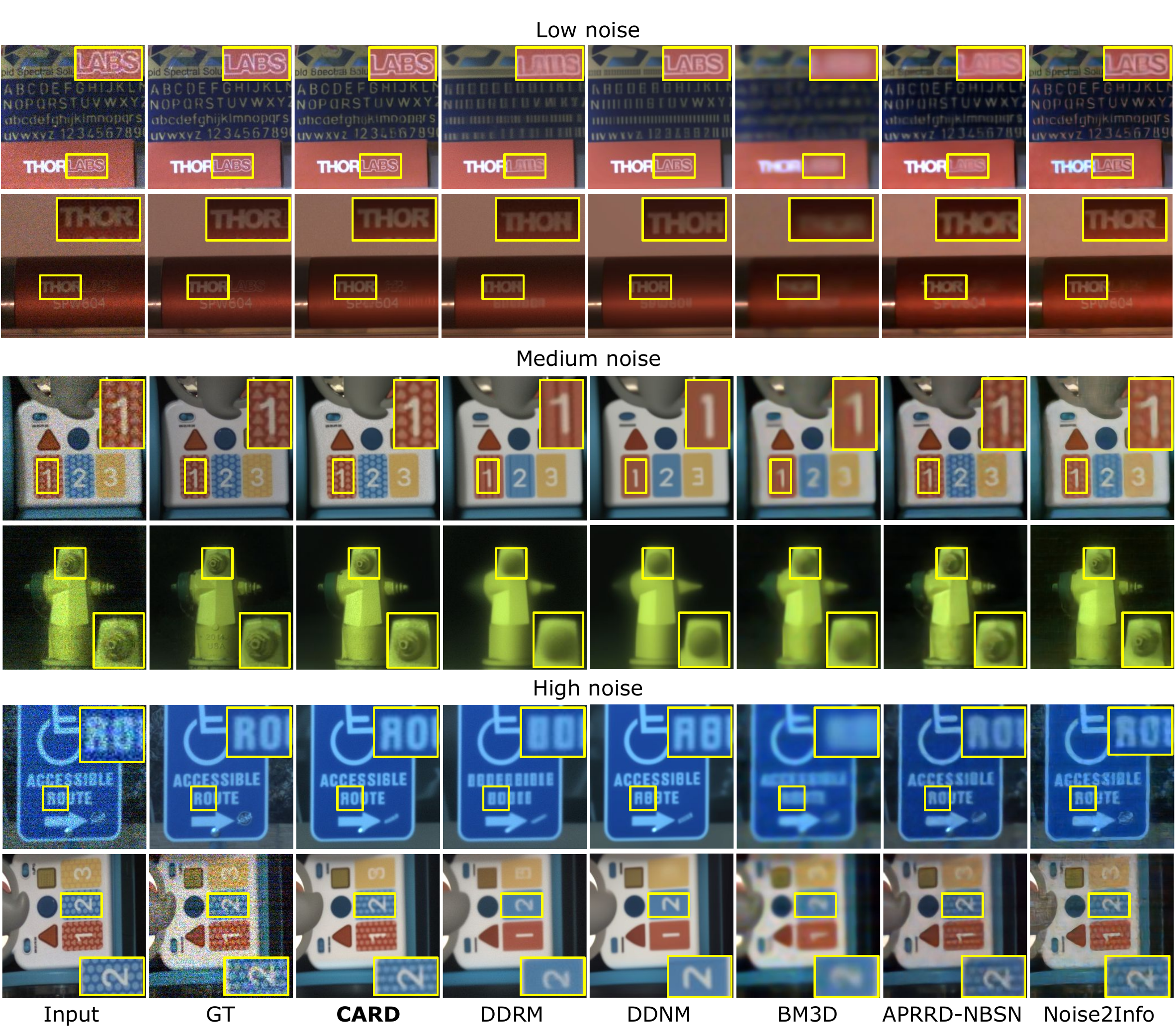}
    \caption{\textbf{Qualitative comparison of denoising on our dataset (CIN-D)} at low, medium, and high noise levels. The boxed regions highlight interesting areas. CARD outperforms other baselines while preserving fine details.}
  \label{fig:supp-deno-real-figure}
\end{figure*}

We evaluate CARD on our dataset CIN-D, which contains indoor and outdoor scenes captured across three noise levels defined by camera gain and exposure. 
\paragraph{Denoising results.}
Table~\ref{tab:deno-real} summarizes the quantitative results. CARD achieves the best PSNR, SSIM, and LPIPS across all noise levels and environments. The improvements are especially significant at medium and high noise, where existing learning-based and diffusion-based methods struggle to preserve structure under strong correlation artifacts. Figure~\ref{fig:supp-deno-real-figure} presents qualitative comparisons. Competing methods either over-smooth fine structures or leave noise artifacts behind, particularly in textured regions and low-light areas. CARD suppresses noise more effectively while retaining high-frequency detail such as textures.

\begin{figure*}[!tt]
  \centering
  \includegraphics[width=0.99\linewidth]{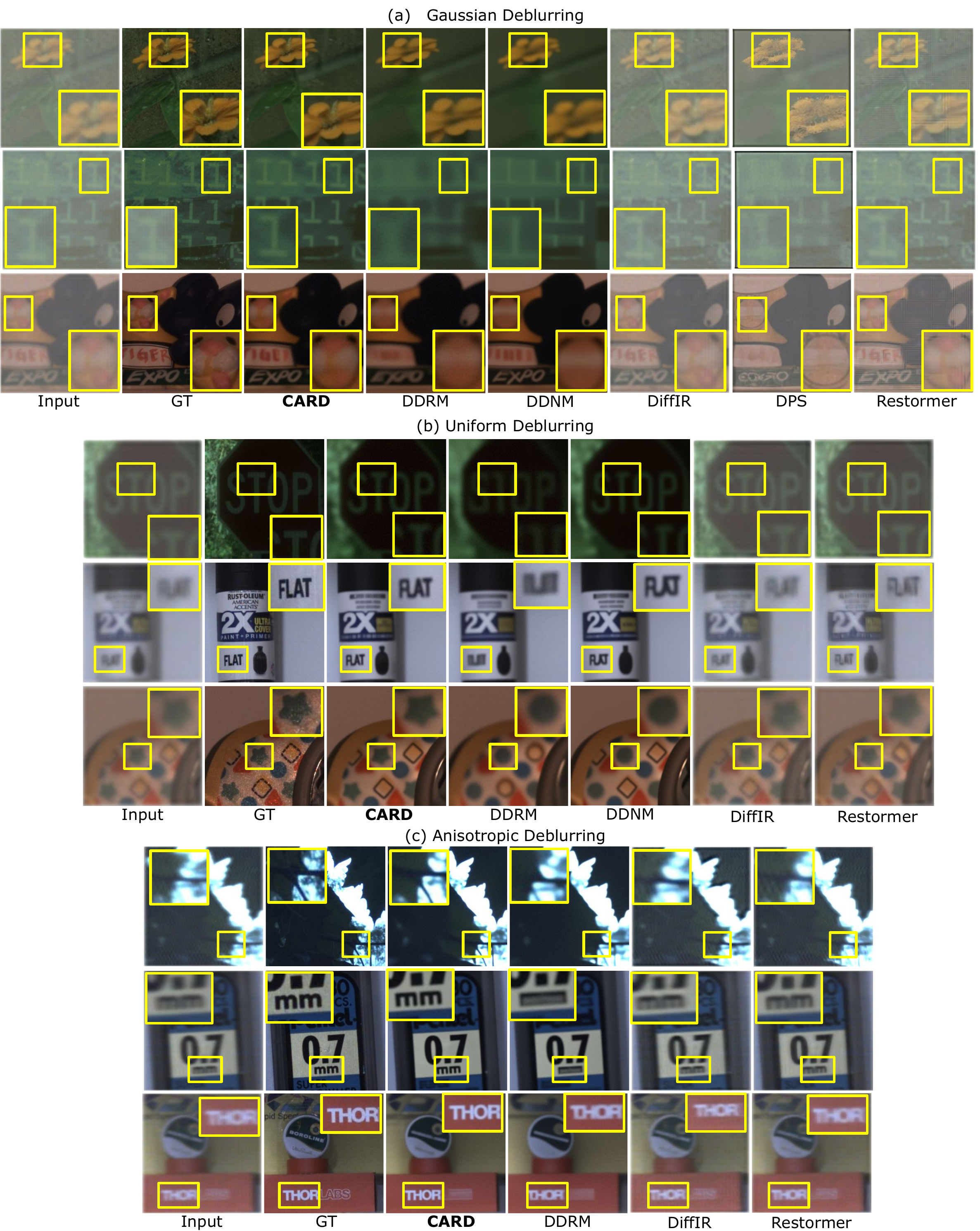}
    \caption{\textbf{Qualitative comparison of Gaussian, Uniform, and Anisotropic deblurring on our dataset (CIN-D)} at low noise level. The boxed regions highlight interesting areas. CARD outperforms other baselines while preserving fine details.}
  \label{fig:supp-deblur-real-figure}
\end{figure*}

\paragraph{Deblurring results.}
We evaluate CARD on deblurring using the forward operators described in Section~\ref{supp:imp_deg_models}. Table~\ref{tab:deblur-cind} shows results at the low-noise setting for both indoor and outdoor scenes. CARD consistently achieves the highest PSNR and SSIM alongside the lowest LPIPS across all three blur types and both scene categories. Competing diffusion-based methods like DDRM and DDNM recover the overall structure but introduce over-smoothing artifacts under correlated noise, as evidenced by their higher LPIPS and lower SSIM scores. Figure~\ref{fig:supp-deblur-real-figure} shows qualitative comparisons. CARD removes noise while preserving fine details such as texts and patterns, whereas other methods either oversmooth or leave structured residuals.

\begin{figure*}[!tt]
  \centering
  \includegraphics[width=0.6\linewidth]{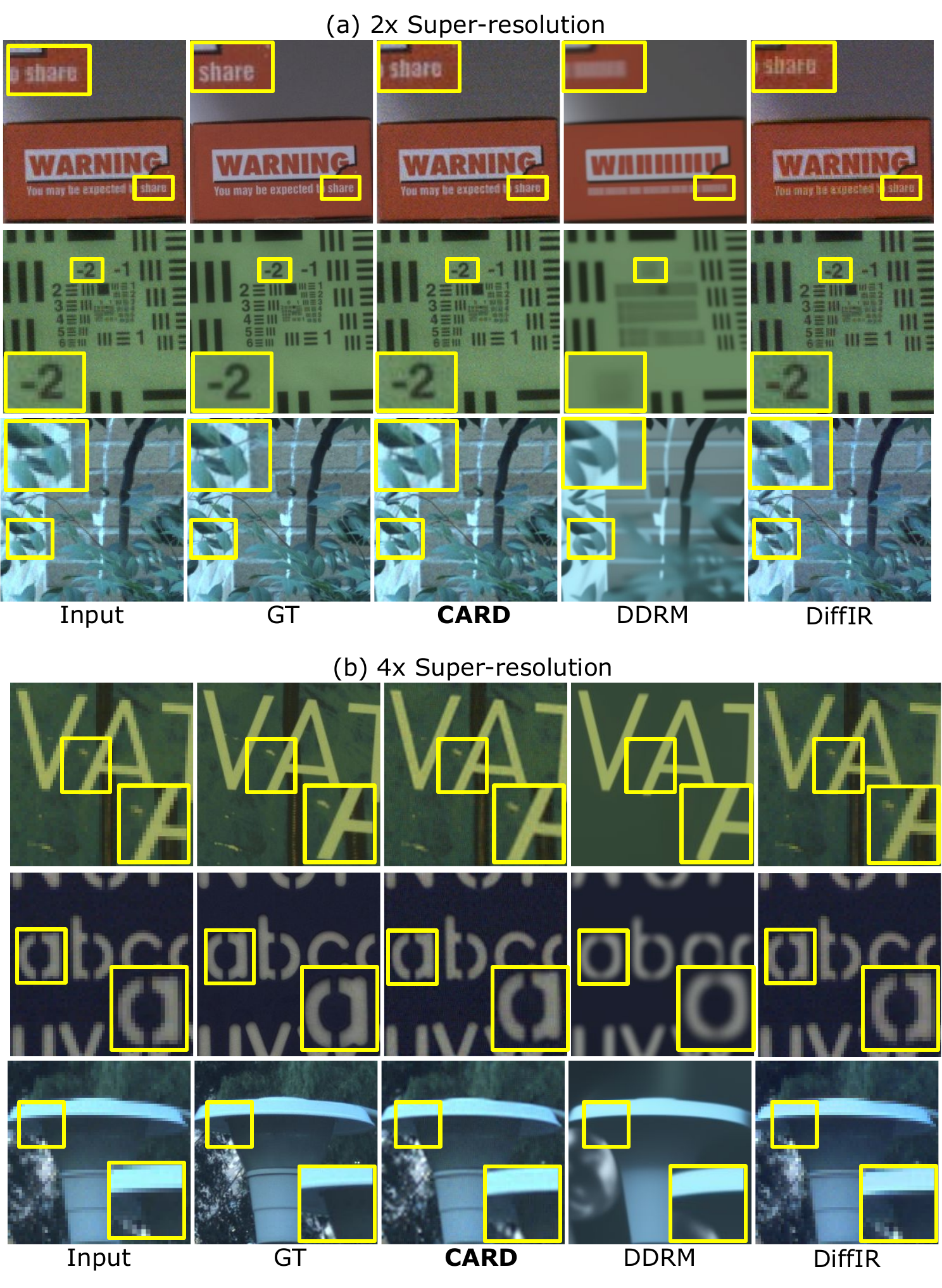}
    \caption{\textbf{Qualitative comparison of $2{\times}$ and $4{\times}$ super-resolution on our dataset (CIN-D)} at low noise level. The boxed regions highlight interesting areas. CARD outperforms other baselines while preserving fine details.}
  \label{fig:supp-sr-real-figure}
\end{figure*}

\paragraph{Super-resolution results.}
We evaluate CARD on $2\times$ and $4\times$ super-resolution using the CIN-D dataset. The forward operators follow the standard degradation models described in Section~\ref{supp:imp_deg_models}. Table~\ref{tab:sr-cind} reports quantitative results at the low-noise setting for both indoor and outdoor scenes. CARD achieves the best performance across all configurations, delivering higher PSNR and SSIM as well as significantly lower LPIPS for both scales. The gains are more significant at $4\times$ upsampling, where correlated noise and aliasing artifacts make the reconstruction particularly challenging. Figure~\ref{fig:supp-sr-real-figure} shows qualitative comparisons. DiffIR and DDRM recover structure but either oversmooth textures or leave noise artifacts.

\begin{table*}[!tt]
\caption{\textbf{Denoising results on our dataset (CIN-D)}, evaluated at low, medium, and high noise levels. We report PSNR/SSIM/LPIPS as P/S/L. Best and second‐best results are marked in bold and underlined, respectively. CARD achieves the best performance at all noise levels.}
\label{tab:deno-real}
\centering
\resizebox{\linewidth}{!}{
\begin{tabular}{@{}lcccccc@{}}
\toprule
\multirow{2}{*}{Model} &
\multicolumn{3}{c}{Indoor Scenes} &
\multicolumn{3}{c}{Outdoor Scenes} \\
& Low (P/S/L)& Med (P/S/L)& High (P/S/L)& Low (P/S/L)& Med (P/S/L)& High (P/S/L)\\
\midrule

DnCNN~\cite{zhang2017dncnn}
& 29.4/0.86/0.29
& 24.5/0.66/0.49
& 25.0/0.47/0.69
& 27.9/0.86/0.29
& 24.7/0.70/0.45
& 24.3/0.50/0.65 \\

Restormer~\cite{zamir2022restormer}
& 16.9/0.47/0.74
& 14.1/0.32/0.84
& 16.1/0.25/0.95
& 17.5/0.48/0.79
& 15.1/0.35/0.87
& 16.4/0.28/0.96 \\

Noise2Info~\cite{wang2023noise2info}
& 12.8/0.65/0.35
& 12.8/0.35/0.87
& 13.5/0.46/0.40
& 19.7/0.71/0.35
& 19.8/0.69/\underline{0.27}
& 20.4/0.60/\underline{0.42} \\

APRRD-BSN~\cite{kim2025apr}
& 14.1/0.68/0.23
& 14.3/0.67/\underline{0.24}
& 14.8/0.65/0.35
& 20.6/0.72/0.32
& 20.5/0.71/\underline{0.27}
& 20.6/0.67/0.43 \\

APRRD-NBSN~\cite{kim2025apr}
& 13.3/0.68/0.26
& 13.3/0.67/0.28
& 13.7/0.65/0.34
& 20.7/0.73/0.30
& 20.2/0.73/0.28
& 20.5/0.70/0.42 \\

PCST~\cite{vaksman2023pcst}
& \underline{38.1/0.93}/0.30
& 27.3/0.67/0.50
& 29.8/0.80/0.34
& \underline{36.3/0.91/0.28}
& 27.3/0.67/0.50
& 29.0/0.76/\underline{0.42} \\

BM3D~\cite{dabov2007bm3d}
& 33.7/0.86/0.31
& 26.8/0.82/0.27
& 29.1/0.78/0.44
& 33.7/0.78/0.43
& \underline{27.9}/0.84/0.35
& 24.2/0.31/0.94 \\

DDNM~\cite{wang2022ddnm}
& 34.2/\underline{0.93}/0.32
& 25.3/\underline{0.86}/0.43
& 24.5/0.60/0.51
& 33.5/\textbf{0.92}/0.35
& 26.7/\underline{0.85}/0.31
& 25.1/0.63/0.50 \\

DDRM~\cite{kawar2022ddrm}
& 36.7/0.89/\underline{0.21}
& \underline{27.7/0.86}/0.25
& \underline{32.3/0.84/0.32}
& 33.9/0.85/0.35
& 27.6/0.81/0.40
& \underline{29.4/0.77}/0.46 \\

\textbf{CARD (ours)}
& \textbf{39.8}/\textbf{0.95}/\textbf{0.17}
& \textbf{27.8}/\textbf{0.88}/\textbf{0.20}
& \textbf{33.1}/\textbf{0.86}/\textbf{0.28}
& \textbf{36.4}/\textbf{0.92}/\textbf{0.22}
& \textbf{28.0}/\textbf{0.87}/\textbf{0.26}
& \textbf{30.0}/\textbf{0.80}/\textbf{0.40} \\

\bottomrule
\end{tabular}}
\end{table*}

\begin{table*}[!tt]
\caption{\textbf{Deblurring results on our dataset (CIN-D)} evaluated at low noise level. We report PSNR/SSIM/LPIPS as P/S/L. Best and second‐best results are marked in bold and underlined, respectively. CARD consistently outperforms all baselines.}
\label{tab:deblur-cind}
\centering
\resizebox{\linewidth}{!}{
\begin{tabular}{@{}lcccccc@{}}
\toprule
\multirow{2}{*}{Model} & \multicolumn{2}{c}{Gaussian Deblur} & \multicolumn{2}{c}{Anisotropic Deblur} & \multicolumn{2}{c}{Uniform Deblur} \\
& Indoor (P/S/L) & Outdoor (P/S/L) & Indoor (P/S/L) & Outdoor (P/S/L) & Indoor (P/S/L) & Outdoor (P/S/L) \\
\midrule

Restormer~\cite{zamir2022restormer} 
& 18.4/0.78/\underline{0.30} 
& 17.9/0.73/\underline{0.39}  
& 26.9/\textbf{0.88}/0.30 
& 21.7/0.78/0.50  
& 22.3/0.82/0.40 
& 21.7/\underline{0.78}/0.50 \\

DPS~\cite{chung2023dps}       
& 18.0/0.74/0.42 & 17.2/0.67/0.54 & -- & -- & -- & -- \\

DiffIR~\cite{xia2023diffir}    
& 19.1/0.81/0.34 
& 18.6/0.76/0.42  
& 25.6/\underline{0.87}/0.33 
& 25.9/\underline{0.82}/\underline{0.38}  
& 21.8/0.82/0.46 
& 21.2/0.77/0.53 \\

DDNM~\cite{wang2022ddnm}     
& 33.2/\underline{0.87}/0.35 
& 32.0/\underline{0.81}/0.40  
& -- 
& -- 
& 32.6/\underline{0.83/0.37} 
& 31.4/0.77/\textbf{0.42} \\

DDRM~\cite{kawar2022ddrm}      
& \underline{34.5}/0.85/0.31 
& \underline{32.1}/0.80/0.44  
& \underline{34.1}/0.85/\underline{0.32} 
& \underline{31.9}/0.79/0.44 
& \underline{33.0}/0.82/\underline{0.37} 
& \underline{30.9}/0.77/\underline{0.48} \\

\textbf{CARD (ours)}   
& \textbf{36.3}/\textbf{0.88}/\textbf{0.23}   
& \textbf{33.9}/\textbf{0.84}/\textbf{0.35}   
& \textbf{35.9}/\textbf{0.88}/\textbf{0.25}    
& \textbf{33.5}/\textbf{0.83}/\textbf{0.37}    
& \textbf{34.9}/\textbf{0.86}/\textbf{0.30}   
& \textbf{32.6}/\textbf{0.81}/\textbf{0.42} \\

\bottomrule
\end{tabular}}
\end{table*}

\begin{table*}[!tt]
\caption{\textbf{Super-resolution results on our dataset (CIN-D)}, across $2\times$ (SR2) and $4\times$ (SR4) upscaling, evaluated at low noise level. We report PSNR/SSIM/LPIPS as P/S/L. Best and second‐best results are marked in bold and underlined, respectively. CARD consistently outperforms all baselines.}
\label{tab:sr-cind}
\centering
\resizebox{0.7\linewidth}{!}{
\begin{tabular}{@{}lcccc@{}}
\toprule
\multirow{2}{*}{Model} & \multicolumn{2}{c}{SR2} & \multicolumn{2}{c}{SR4} \\
& Indoor (P/S/L) & Outdoor (P/S/L) & Indoor (P/S/L) & Outdoor (P/S/L) \\
\midrule
DiffIR~\cite{xia2023diffir}                 & \underline{37.5/0.92/0.19} & \underline{35.6/0.91/0.18}               & \underline{35.5/0.89/0.24}              & \underline{33.7/0.87/0.29}               \\
DDRM~\cite{kawar2022ddrm}                   & 35.7/0.92/0.28 & 32.9/0.87/0.41  & 32.5/0.88/0.39 & 30.7/0.84/0.49  \\
\textbf{CARD (ours)}   & \textbf{39.5/0.94/0.14} & \textbf{37.4/0.93/0.12}  & \textbf{37.2/0.91/0.20} & \textbf{35.1/0.89/0.22}  \\
\bottomrule
\end{tabular}}
\end{table*}

\end{document}